\documentclass{article}

\PassOptionsToPackage{numbers,sort,compress}{natbib}

\usepackage[preprint]{neurips_2020} %arXiv

\usepackage[utf8]{inputenc} % allow utf-8 input
\usepackage[T1]{fontenc}    % use 8-bit T1 fonts
\usepackage{hyperref}       % hyperlinks
\usepackage{url}            % simple URL typesetting
\usepackage{booktabs}       % professional-quality tables
\usepackage{amsfonts}       % blackboard math symbols
\usepackage{nicefrac}       % compact symbols for 1/2, etc.
\usepackage{microtype}      % microtypography

\usepackage{graphicx}
\usepackage{amsmath}
\usepackage{longtable}

\usepackage{cleveref}
\Crefname{equation}{Equation}{Equations}
\crefname{equation}{Equation}{Equations}
\Crefname{figure}{Figure}{Figures}
\crefname{figure}{Figure}{Figures}
\Crefname{table}{Table}{Tables}
\crefname{table}{Table}{Tables}
\Crefname{section}{Section}{Sections}
\crefname{section}{Section}{Sections}
\Crefname{appendix}{Appendix}{Appendices}
\crefname{appendix}{Appendix}{Appendices}

\newcommand{\y}{y}

\newcommand{\z}{\mathcal{Z}}

\newcommand{\nl}{L}

\newcommand{\sigy}{\sigma_{\y}}
\newcommand{\np}{n_{\mathrm{pars}}}

\newcommand{\e}{E}
\newcommand{\ga}{\alpha_i}
\newcommand{\gb}{\beta_i}
\newcommand{\PolyChord}{\texttt{PolyChord}}
\newcommand{\MultiNest}{\texttt{MultiNest}}
\DeclareMathOperator{\relu}{ReLU}

\title{Compromise-free Bayesian neural networks}

\author{%
  Kamran Javid, \hfill Will Handley, \hfill Mike Hobson \& \hfill Anthony Lasenby \\
  Cavendish Laboratory (Astrophysics Group) \\
  University of Cambridge, UK, CB3 9HE \\
  \texttt{kj316@cam.ac.uk}, \texttt{wh260@cam.ac.uk}
}

\begin{document}

\maketitle

\begin{abstract}
We conduct a thorough analysis of the relationship between the out-of-sample performance and the Bayesian evidence (marginal likelihood) of Bayesian neural networks (BNNs), as well as looking at the performance of ensembles of BNNs, both using the Boston housing dataset. Using the state-of-the-art in nested sampling, we numerically sample the full (non-Gaussian and multimodal) network posterior and obtain numerical estimates of the Bayesian evidence, considering network models with up to 156 trainable parameters. The networks have between zero and four hidden layers, either $\tanh$ or $\relu$ activation functions, and with and without hierarchical priors. The ensembles of BNNs are obtained by determining the posterior distribution over networks, from the posterior samples of individual BNNs re-weighted by the associated Bayesian evidence values. There is good correlation between out-of-sample performance and evidence, as well as a remarkable symmetry between the evidence versus model size and out-of-sample performance versus model size planes. Networks with $\relu$ activation functions have consistently higher evidences than those with $\tanh$ functions, and this is reflected in their out-of-sample performance. Ensembling over architectures acts to further improve performance relative to the individual BNNs.
\end{abstract}
\section{Introduction}

%TODO can possibly drop this for space
In an age where machine learning models are being applied to scenarios where the associated decisions can have significant consequences, quantifying model uncertainty is becoming more and more crucial \citep{krzywinski2013points, ghahramani2015probabilistic}. Bayesian neural networks (BNNs) are one example of a model which provides its own uncertainty quantification, and have gained popularity in-part due to the successes of the conventional backward propagation trained neural networks \citep{rumelhart1985learning}.

BNNs have a history stretching back almost as far as research on traditional neural networks (TNNs), with \citep{denker1987large, tishby1989consistent, buntine1991bayesian, denker1991transforming} laying the foundations for the application of BNNs. Major breakthroughs occurred thanks to the work of \citet{mackay1994bayesian} and his success with BNNs in prediction competitions. Prior to this, \citeauthor{mackay1992bayesian} published several papers \citep{mackay1992bayesian, mackay1992practical, mackay1992evidence} detailing his methods and highlighting several important aspects of BNNs. He trained the networks by using quadratic approximations of the posteriors, type II maximum likelihood estimation \citep{rasmussen2003gaussian}, and incorporated low-level hierarchical Bayesian inference into his modelling to learn the prior distribution variances. From this he obtained estimates for the Bayesian evidence (also known as the marginal likelihood), and found a correlation between the evidence and the BNNs' ability to make predictions on out-of-sample data well. 
\citet{neal1992bayesian, neal1993bayesian} focused on improving the predictive power of BNNs, by relaxing the Gaussian approximation by instead sampling the posterior using Hamiltonian Markov chain Monte Carlo (MCMC) methods. He also incorporated more complex hierarchical Bayesian modelling into his methods by using Gibbs sampling to sample the variances associated with both the priors and likelihood. He found that predictions with these BNNs consistently outperformed the equivalently sized networks trained using backward propagation techniques. More recently \citep{freitas2000sequential, de2003bayesian, andrieu2003introduction} used reversible jump MCMC and sequential MC methods to train BNNs and do model selection without calculating Bayesian evidences, by parameterising the different networks as a random variable, and sampling from the resultant posterior. In general they found that their methods produce networks which perform well, but were much slower to train than expectation maximisation-based networks. Indeed, methods which are more efficient in the training of networks have gained popularity in the recent years, due to the success of deep learning. Two of the most commonly used methods for training scalable networks are variational inference \citep{hinton1993keeping, barber1998ensemble, graves2011practical} and dropout training \citep{gal2016dropout}. The latter has proved to be particularly popular, due to the fact that dropout BNNs can be trained using standard backward propagation techniques, and can be applied to recurrent and convolutional networks \citep{gal2016theoretically, kendall2017uncertainties}.

In this paper we present what we refer to as {\em compromise-free\/} BNNs as a proof-of-concept idea for training BNNs, conditional on computational resources. As in \citet{higson2018bayesian}, no assumption is made about the functional form of the posteriors, and we numerically sample the full posterior distributions using the nested sampling algorithm \PolyChord{} \citep{handley2015polychord1, handley2015polychord2}\footnote{\url{https://github.com/PolyChord/PolyChordLite}} to train the BNNs. Furthermore we obtain numerical estimates for the Bayesian evidences associated with each BNN, with which we analyse the relationship between the evidence and out-of-sample performance as originally performed approximately in \citet{mackay1992practical}. We consider a wide array of different networks, with between zero and four hidden layers, $\tanh$ or $\relu$ activation functions, and varying complexities of hierarchical priors following \citep{mackay1992practical, neal2012bayesian}, to obtain a thorough understanding of the evidence and out-of-sample performance through various cross sections of the BNN architecture space.

%TODO Could make this shorter
 Similarly to \citet{de2003bayesian} we look at the posterior over networks as a form of model selection. However in our case, the posteriors are obtained from the samples of the individual network posteriors re-weighted according to the evidences associated with a given run. We then marginalise over these network posteriors to obtain predictions from ensembles of BNNs. A preliminary analysis along these lines was conducted by \citet{higson2018bayesian}, who also explored using an adaptive method akin to \citet{de2003bayesian}. The adaptive approach has the feature of severely undersampling less preferred models and taking (often substantially) less time than sampling models individually, and will be explored in a future work.

\section{Background}\label{s:background}
\subsection{Neural networks}\label{s:nn}
A fully connected multi-layer perceptron (MLP) neural network is parameterised by network weights $w$ and biases $b$, and can be represented recursively using intermediate variables $z$ as
\begin{equation}
    f = z^{(\nl)}, \qquad
    z^{(\ell)}_i =  g_i^{(\ell)}(b_i^{(\ell)} + \textstyle\sum_{k=1}^{l_i} w_{ik}^{(\ell)} z_{k}^{(\ell-1)}),\qquad
    z^{(0)} = x,
    \label{e:nn}
\end{equation}
where $g_i^{(\ell)}$ are {\em activation functions}, which we take to be either $\tanh$ ($g(x) = \tanh x $) or $\relu$ ($g(x) = \max\{x,0\}$) in the hidden layers, with a linear activation in the output layer $\ell=L$. Such networks are represented graphically in \cref{f:nn}. The activation functions $g$, number of layers $\nl$, and the number of nodes within each layer $l_i$ together determine the {\em architecture\/} of the network. 
The network parameters $\theta=(b,w)$ are trained in a supervised fashion on a set of example input data paired with the corresponding outputs via a misfit function and regularisation term
\begin{equation}\label{e:chi2}
    \theta_* = \min\limits_{\theta} \lambda_{m} \chi^2_\mathrm{train}(\theta) + \lambda_{r} R(\theta), \qquad \chi^2_\mathrm{train}(\theta) = \textstyle\sum_{i\in\mathrm{train}} |y^{(i)} - f(x^{(i)};\theta)|^2,
\end{equation}
where $\lambda_{m}$ and $\lambda_{r}$ are hyperparameters dictating the relative weighting of the misfit versus the regularisation in the optimisation.\footnote{It should be noted that in a minimisation context, there is a redundancy in including two regularisation parameters $\lambda_m$ and $\lambda_r$, so people usually without loss of generality set $\lambda_m=1$. In a Bayesian context this redundancy is removed, as the posterior is composed of likelihood and prior terms whose widths are each controlled by a separate regularisation parameter. We thus follow \citet{mackay1994bayesian}, retaining both $\lambda_m$ and $\lambda_r$.} 

There are two key issues that immediately arise from the traditional approach. First, the network gives no indication of the confidence in its prediction $y_\mathrm{pred} \equiv f(x_\mathrm{new};\theta_*)$ on unseen inputs $x_\mathrm{new}$. It would be preferable if the network were to provide an error bar $y_\mathrm{pred}\pm\sigma_{y, \mathrm{pred}}$ for its confidence, and that this error bar should become larger as the network extrapolates beyond the domain of the original training data. Second, there is little guidance from the formalism above as to how to choose the architecture of the network in \cref{e:nn} and the hyperparameters in \cref{e:chi2}. Networks that are too large may overfit the data, while networks too small/simple will likely underfit. The most common method of finding a `happy medium' is through the use of cross validation \citep{stone1977asymptotics, efron1979computers, li1985stein, snoek2012practical}, but searching the associated hyperparameter space can be time consuming and ad-hoc.%\footnote{One also runs the risk that if one tries too many different networks one brings the testing data ``back into sample'' manually by performing what is effectively an optimisation over networks and testing data.}

\subsection{Bayesian neural networks} \label{s:bnn}
\begin{figure}
    \centerline{\includegraphics[width=\textwidth]{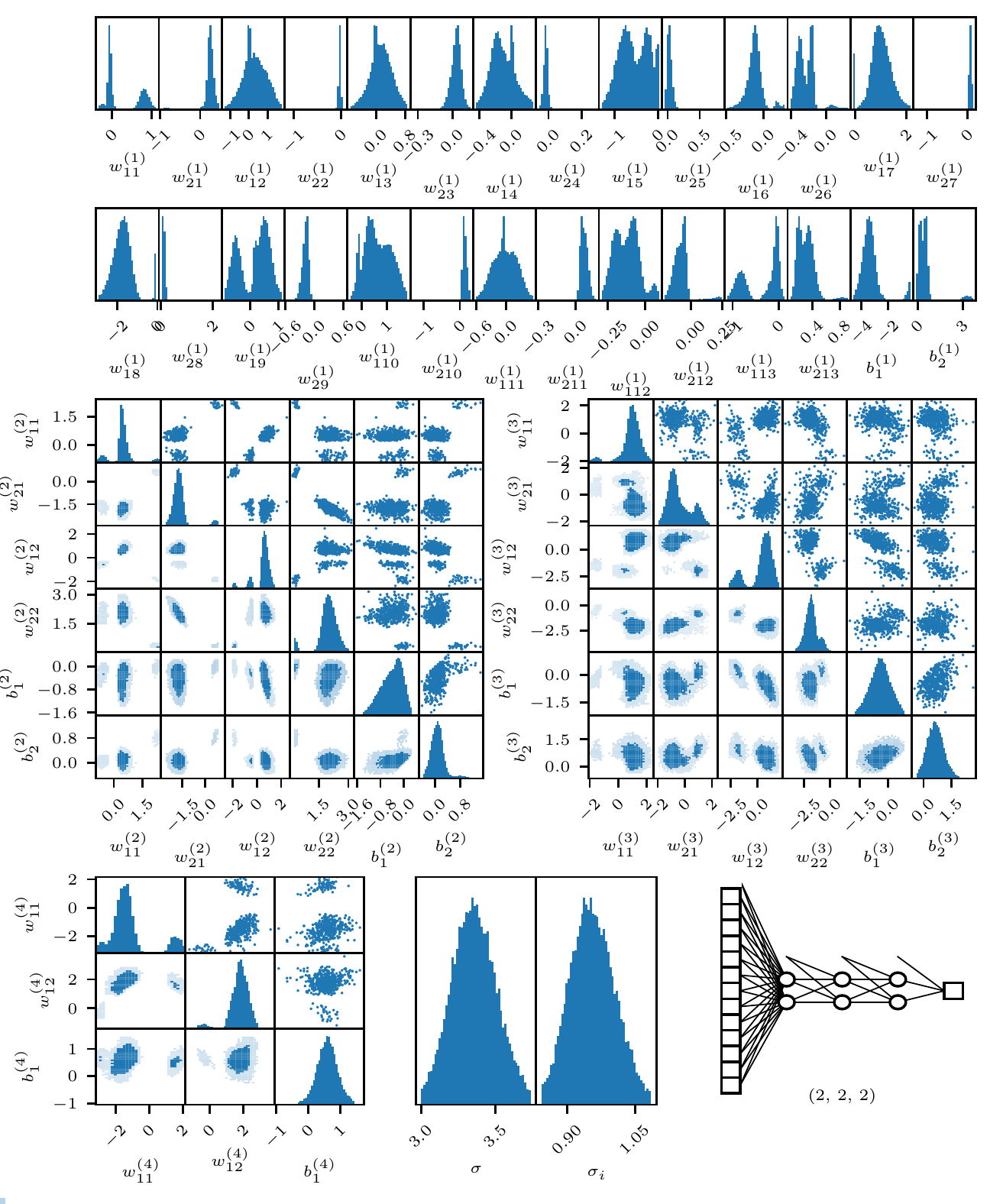}}
    \caption{Example posterior produced by \PolyChord{}~\citep{handley2015polychord1,handley2015polychord2} for the neural network shown in the bottom right corner with ``single'' granularity hyperpriors (\cref{s:further_bayes}). The full 45-dimensional posterior is sampled numerically. In the top panels the 1-dimensional marginal distributions for the network weights $w$ and biases $b$ in the first layer are shown. The three square plots show the pairwise 2-dimensional marginal posteriors for the next three layers, with 1-d marginals on the diagonals, representative samples drawn from the posterior in the upper triangle and histograms of the posterior in the lower triangle. The bottom centre plots show the posterior on the {\em a-priori\/} unknown Gaussian noise level $\sigma$ in the likelihood and single prior hyperparameter $\sigma_1$. It should be noted that the posteriors are highly non-Gaussian and multi-modal, necessitating the use of a full compromise-free sampler. Plot created under \texttt{anesthetic}~\citep{2019JOSS....4.1414H}.
        \label{f:posteriors}
}
\end{figure}

The Bayesian approach to neural networks aims to ameliorate the two difficulties discussed above in \cref{s:nn} by {\em sampling\/} the parameter space rather than {\em optimising\/} over it. Here we consider the misfit function $\chi^2_\mathrm{train}(\theta)$ as part of an independent Gaussian likelihood
\begin{equation}
    P(D_\mathrm{train}|\theta,\mathcal{M}) = \prod_{i\in\mathrm{train}} \frac{1}{\sqrt{2\pi}\sigma}\exp{\left(-\frac{(y^{(i)}-f(x^{(i)};\theta))^2}{2\sigma^2}\right)} = \frac{\exp\left(-\frac{\chi^2_\mathrm{train}(\theta)}{2\sigma^2}\right) }{{(\sqrt{2\pi}\sigma)}^{N_\mathrm{train}}},
    \label{e:like}
\end{equation}
where $D_\mathrm{train}$ are the training data, $\sigma^2$ is a misfit variance (which plays a similar role to the parameter $\lambda_m$ in \cref{e:chi2}) and $\mathcal{M}$ is the network architecture (or Bayesian model). The likelihood $\mathcal{L}\equiv P(D_\mathrm{train}|\theta,\mathcal{M})$ can be related to a posterior $\mathcal{P}$ on the parameters $\theta$ using a prior measure on the network parameter space $P(\theta|\mathcal{M})\equiv\pi$ (which draws parallels with $\lambda_r R(\theta)$~\citep{higson2018bayesian}) via Bayes theorem
\begin{equation}
    \mathcal{P}\equiv P(\theta|D_\mathrm{train},\mathcal{M}) = \frac{P(D_\mathrm{train}|\theta,\mathcal{M}) P(\theta|\mathcal{M})}{P(D_\mathrm{train}|\mathcal{M})} \equiv \frac{\mathcal{L}\pi}{\mathcal{Z}}, \qquad
    \mathcal{Z} = \int \mathcal{L}\pi \mathrm{d}\theta.
    \label{e:bayes}
\end{equation}
The process of determining the posterior is termed {\em parameter estimation}.
Instead of having a single best-fit set of network parameters $\theta_*$, one now has a distribution over $\theta$ (\cref{f:posteriors}).
This quantification of error in the posterior may be forwarded onto a distribution on the predictions $y_\mathrm{pred}$, from unseen inputs $x_\mathrm{new}$, which may be summarised by a mean $\hat{y}_\mathrm{pred}$ and an error bar $\sigma_{y,\mathrm{pred}}^2$.  

The other critical quantity $\mathcal{Z}$ in \cref{e:bayes} is termed the Bayesian evidence (also known as the marginal likelihood), and is computed from the likelihood and prior as a normalisation constant. The evidence is critical in the upper level of Bayesian inference, termed {\em model comparison}, whereby one's confidence in the network as supported by the data is given by
\begin{equation}
    P(\mathcal{M}|D_\mathrm{train}) = \frac{P(D_\mathrm{train}|\mathcal{M})P(\mathcal{M})}{P(D_\mathrm{train})} = \frac{\mathcal{Z}_\mathcal{M}P(\mathcal{M})}{\sum_m \mathcal{Z}_m P(m)}.
\end{equation}
In the above posterior over models, $m$ is a categorical variable ranging over all architectures considered, and $P(m)$ is the assigned prior probability to each network (typically taken to be uniform over all $m$). The evidence is therefore a measure of the quality of a network as viewed by the data and can be used to compare architectures and marginalise over models.

The Bayesian evidence $\mathcal{Z}$ in \cref{e:bayes} is the average of the likelihood function over the parameter space, weighted by the prior distribution. A larger parameter space, either in the form of higher dimensionality or a larger domain results in a lower evidence value, all other things being equal. Thus the evidence automatically implements Occam's razor: when you have two competing theories that make similar predictions, the one with fewer active (i.e.\ constrained) parameters should be preferred. 

This naturally has useful implications in the context of machine learning: for two models which fit the training data equally well, one would expect the simpler model (i.e. the one with the higher Bayesian evidence) to generalise to out-of-sample data better, due to it overfitting the training data less. Thus one can postulate that the Bayesian evidence can be used as a proxy for out-of-sample data performance of a model relative to alternative models. If shown to be true in practice more generally, this has wide-reaching implications for machine learning, as an orthogonal measure of performance generalisation from training data alone might allow less data to be sacrificed to testing sets, and a more robust test of performance on completely unseen data.

\citet{mackay1992practical} historically found a good correlation between $\z$ and generalisation ability for small neural networks applied to regression problems. Furthermore \citeauthor{mackay1992practical} found that for models where this correlation did not exist, the models generally performed poorly on test set data, but when they were improved in some way \footnote{In the particular example presented, \citet{mackay1992practical} improved performance by increasing the granularity of the variable prior hyperparameters.} then the correlation was found to exist. Thus in this instance the correlation (or lack thereof) between the evidence and out of sample performance can also be interpreted as a tool for determining when a model's performance on out-of-sample data can be improved. One of the key results of this paper is the demonstration of the correlation between $\z$ and generalisation ability to be true in the compromise-free setting as well.

%TODO can possibly make this shorter
A further use of the evidence $\z$ is to perform Bayesian ensembling over networks/models. Taking a uniform prior over all networks, samples from the full ensemble distribution over different networks can be generated by re-weighting so that the posterior mass of each individual network is proportional to its evidence. This generates a set of samples from the full joint distribution. In the context of MLP neural networks, different models can take several forms, including different numbers of layers, different numbers of nodes within the layers, different activation functions, and in our case, even different granularities (see \cref{s:methods}) on the hierarchical priors. The beauty of using the full joint distribution is it allows the data to decide which models are most important in the fitting process, arguably putting less onus on the user as they now only have to decide on a suite of models to choose from, and the associated prior probabilities of these models. Supplementary detail may be found in \cref{s:further_bayes}.

\section{Methodology}\label{s:methods}

\textbf{Data:} For this paper we focus on the Boston housing dataset\footnote{\url{https://archive.ics.uci.edu/ml/machine-learning-databases/housing/}} as its small sample size is appropriate for a compromise-free Bayesian treatment. The dataset consists of 506 continuously variable records, with 13 inputs (features) and one output. We linearly whiten the input and output variables of this regression problem so that they have zero mean and unit variance.
For each analysis we split the entire data in half so that both the training and test sets contain $n_\mathrm{test}=n_\mathrm{train}=253$ records. We evaluate the test set performance by considering the mean squared error $E=\chi^2_\mathrm{test}/{n_\mathrm{test}}$ between $y_{\mathrm{test}}$ and the mean BNN prediction $\widehat{y}_\mathrm{new}$, as well as the error on $E$ derived from $\sigma_{y, \mathrm{pred}}$. 

For training purposes we repeat the analysis with ten different random splits of the training/test data, so for a given setup we train a BNN on ten different training sets and measure their performance on the ten corresponding test sets. For these ten different data splits, we look at the average value of the mean squared errors on the test sets, and also take the mean value of $\z$ obtained from the ten analyses. The values quoted in the rest of this analysis are the logarithm of the values of these average values of the evidence, and the average values of the mean squared errors.

\textbf{Models:} \cref{f:nn} details the neural network architectures we consider alongside a network with no hidden layer (i.e. Bayesian linear regression).  
For each architecture, we first consider networks with either a $\tanh$ or $\relu$ activation function for all hidden layers, with Gaussian prior and likelihood widths fixed to $\sigma_i=\sigma=1$. 
For the $\tanh$ activation functions, we also consider the impact of hierarchical priors as discussed in \cref{s:further_bayes}.\footnote{This is analogous to letting the data decide on the values of $\lambda_m$ and $\lambda_r$} In this case we also allow the likelihood precision $\tau \equiv \sigma^{-2}$ to vary by setting a Gamma prior with $\alpha=\beta=1$. 
In the terminology of Bayesian statistics we shall refer to each combination of architecture, activation function and prior as a {\em model}.

For the hierarchical priors, three different variants were considered, corresponding to three different levels of granularity on the priors: single, layer \citep[both used in][]{mackay1992practical} and input size \citep[similar to the automatic relevance determination method in][]{neal2012bayesian}. Single granularity has one hyperparameter controlling the standard deviations on the priors of all weights and biases in the network. Layer granularity has two hyperparameters per layer (one for the bias, one for the weights in each layer). For input size granularity models, the number of hyperparameters for each layer depends on the number of inputs to that layer. Following \citet{neal2012bayesian} we used zero mean Gaussian distributions with an ordering enforced to prevent weight space degeneracy \citep{goodfellow2016deep} \citep[as discussed in][]{handley2019bayesian, buscicchio2019label} for the priors and Gamma distributions for the hyperpriors. For layer and input size random variable hyperparameter models we scaled the Gamma distribution hyperparameters so that the prior over functions converges to Gaussian processes in the limit of infinitely wide networks, as discussed in \citet{neal2012bayesian}. 

\textbf{Training:} To obtain estimates of the BNN posterior distributions and Bayesian evidences we use the \PolyChord{} algorithm \citep{handley2015polychord1, handley2015polychord2}, a high-dimensional, high-performance implementation of nested sampling \citep{skilling2006nested}. We run with 1,000 live points $n_\mathrm{live}$ and the number of repeats $n_\mathrm{repeats}$ set to 5 $\times$ the dimensionality of the parameter space (which vary between $14$ and $156$ dimensions, see \cref{t:non_combined_averages}). 
We followed the recommended procedure of checking our results are stable with respect to varying $n_\mathrm{repeats}$. An example posterior produced by \PolyChord{} is plotted under \texttt{anesthetic}~\citep{2019JOSS....4.1414H} in \cref{f:posteriors}, which shows the compromise-free posterior as highly non-Gaussian, with nonlinear correlation structure and significant multimodality. A naive optimiser applied to such a training scenario would likely be unable to reveal such information.

\textbf{Computing:} Our longest runs (those with the biggest networks and most complex hierarchical priors) took up to 12 hours using the multithreaded Eigen\footnote{\url{http://eigen.tuxfamily.org/index.php?title=Main_Page}} C++ library to implement these BNNs, computed on Intel Xeon Skylake 6142 processors (16-core 2.6GHz, 192GB RAM)\footnote{\url{https://www.hpc.cam.ac.uk/systems/peta-4}}. \PolyChord{} may be MPI parallelised up to the number of live points (i.e. 1,000), but we opted to trivially parallelise over separate network runs. We also experimented with serial tensorflow-gpu code on a Nvidia P100 GPU 16GB GPU \footnote{\url{https://www.hpc.cam.ac.uk/systems/wilkes-2}}, but found this to be slower than the CPU code described above, due to the fact that we were running small networks on small datasets, so the transferring in and out of GPU memory overhead for subsequent samples outweighed the GPU matrix manipulation gains. The code used to conduct these experiments is publicly available on GitHub\footnote{\url{https://github.com/SuperKam91/bnn}}.

\begin{figure}
    \centerline{ \includegraphics{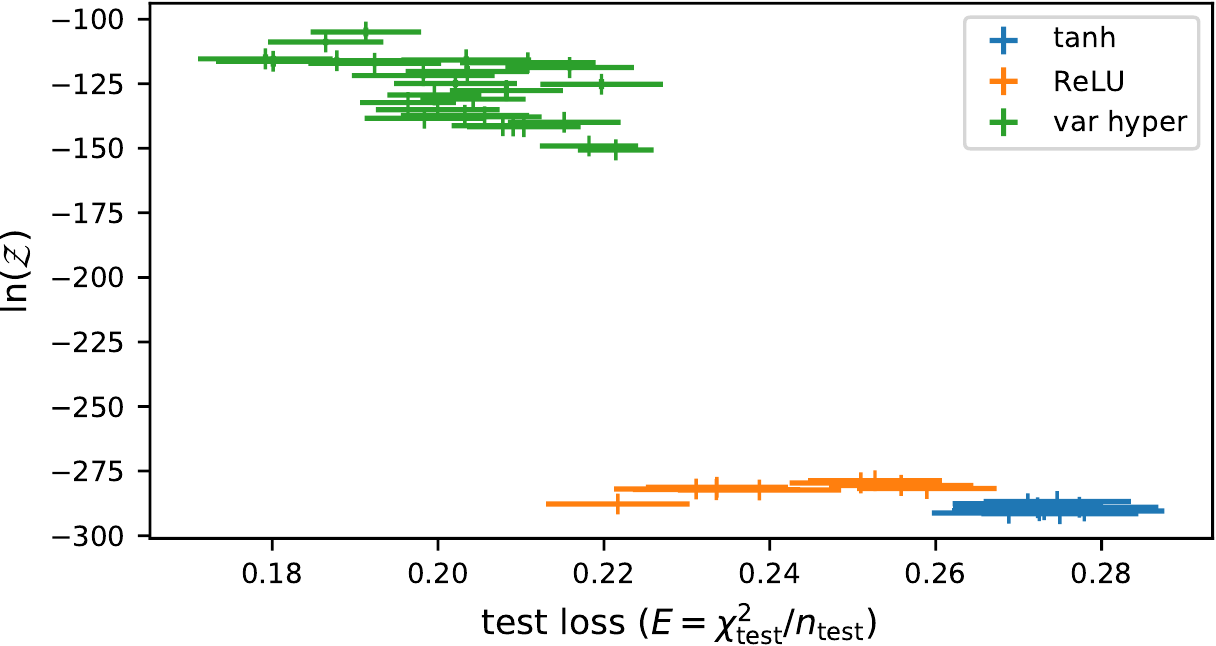}}
  \caption{Log Bayesian evidence versus test loss values averaged over the ten different data randomisations, for the BNNs with $\tanh$ activation functions (blue), $\relu$ activations (orange), and variable hyperparameter models with $\tanh$ activations (green).  }
\label{f:non_comb_mean_results}
\end{figure}

\section{Results}\label{s:results}

%TODO possibly split into subsections
In total we trained 49 different model combinations of architecture, activation function and prior (\cref{t:non_combined_averages}), and for each we performed 10 different randomisations of the training--test split (in each case 50\% of the data is used for training and the remaining 50\% is used for testing), and took averages of these 10 results. We examined correlations between the test loss (mean squared error of the BNN mean estimates on the test data), the Bayesian evidences, and the size of the parameter space dimensionalities for the different models. A more detailed discussion of results with additional figures and tables may be found in \cref{s:detailed_results}.

Our headline results are shown as a plot in the test-set performance--evidence plane in \cref{f:non_comb_mean_results}. Across all of the models there are three distinct clusters: models which used fixed prior/likelihood variances split into $\tanh$ and $\relu$ clusters, and models which used hierarchical priors (which all used $\tanh$ hidden activations). In all cases for a given architecture, the latter had higher evidence values and better performance, adhering to the trend found in \citet{mackay1992practical}.
	
For models with fixed prior/likelihood variances, the models which used the $\relu$ activation function consistently outperformed the equivalent architectures with $\tanh$ activations. 
This is a somewhat surprising result (further highlighted in \cref{f:non_comb_tanh_relu_mean_results}) since $\tanh$ is a more non-linear function, one may expect it to perform better than $\relu$ for the small networks considered here. In traditional neural network training, two of the key reasons why $\relu$ is a popular choice are that: 1) its derivative is fast and easy to calculate which is crucial for backward propagation training and 2) the fact that non-positive activations are shut out (their value and derivative are both zero) means that a side-effect of using $\relu$ is that it provides a form of regularisation, which can be helpful in large networks. Neither of these considerations are applicable to the BNNs we consider here, however, since \PolyChord{} does not use derivative information, and only small networks are considered. One potential benefit of using $\relu$ for these BNNs is that the function does not saturate for input values large in magnitude as $\tanh$ does. Regardless of why $\relu$ so consistently outperforms the $\tanh$ models, the evidence clearly picks up on its superiority.

\begin{figure}
    \centerline{\includegraphics{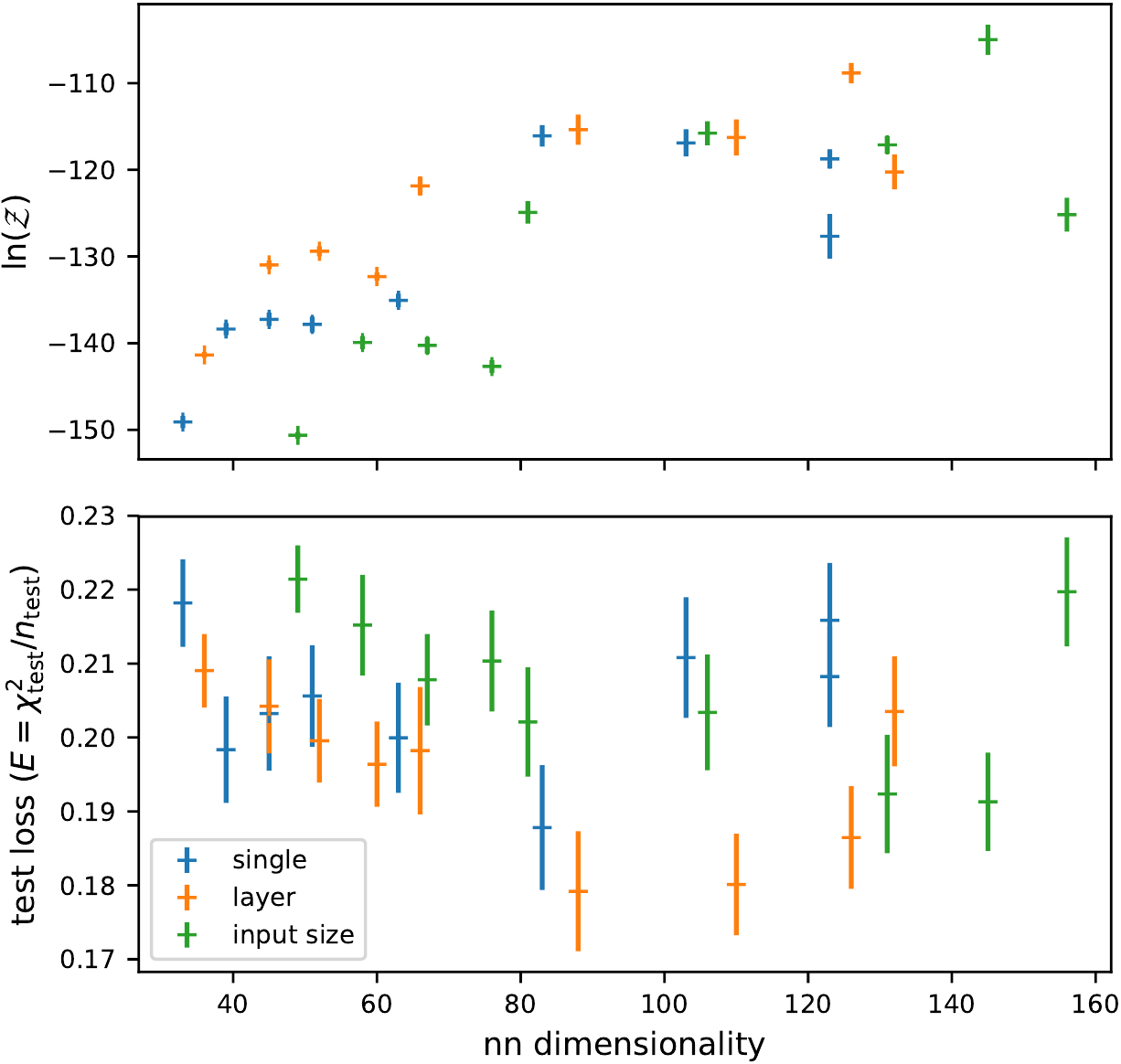}}
    \caption{Log Bayesian evidence (top plot) and test loss values (bottom plot) versus BNN model dimensionality focussing on the cases with variable hyperparameters with single (blue points), layer (orange points), or input size granularity. 
    Note that the mapping between architecture and dimensionality is not one-to-one, as some models by chance have the same dimensionality. In these cases, the evidence may still be used to select between architectures of the same dimensionality. For clarity, reading right-to-left and then top-to-bottom the network architectures are (2) (2,2) (2,2,2) (2,2,2,2), (4), (4,4), (4,4,4), (4,4,4,4), (8). 
}
\label{f:non_comb_sh_lh_ih_sv_mean_results_mean_Z_vs_dims_mean_loss_vs_dims}
\end{figure}

We also investigated the effect of using hierarchical priors with a range of degrees of complexity/granularity (single, input and layer: \cref{s:stochasticresults}). At single (course) granularity, there is a definite correlation between test set accuracy and Bayesian evidence (\cref{f:non_comb_sh_lh_ih_sv_mean_results}).  
For the same subset of models, there appears to be a Bayesian cliff present when considering the evidence as a function of parameter space dimensionality \citep{mackay1992practical}, as well as a striking symmetry between the evidence--dimensionality and the test set performance--dimensionality planes (\cref{f:non_comb_sh_lh_ih_sv_mean_results_mean_Z_vs_dims_mean_loss_vs_dims}). 
A stronger correlation between performance and evidence is present for the layer random variable prior hyperparameter models. The symmetry mentioned previously is also present, but in this case the Bayesian cliff is more of a plateau. The same can be said for the models with input size granularity.
For layer granularity models the test set performance was better than the equivalent models with single granularity in most cases. The evidence values were also higher for the former in general, indicating that it correctly captured the superior performance. But perhaps surprisingly, the input size granularity models consistently underperform the layer granularity models, in contrast to \citep{neal2012bayesian, javid201921cm}. Further investigation into the training and test set losses suggests that the input size models are on average shutting off nodes important for the test data, more than the layer granularity models are, while performing similarly well on the training data.

The jump in performance and evidence, and increase in correlation between the two, associated with switching from fixed to variable prior/likelihood variances was also found in \citet{mackay1992practical}, but in effectively switching from single to layer granularity models. \citeauthor{mackay1992practical} attributed this to the idea that when a correlation between evidence and performance was not present, then the model could be improved in some way, and that the improved model showed this correlation more.

We finally combined the predictions of an ensemble of BNNs by considering the posterior distribution over the corresponding models, parameterised by a categorical variable representing a given BNN (\cref{t:combined_averages}). The corresponding posteriors were obtained from the samples of the individual model posteriors which are re-weighted according to the evidences associated with a given run, and the ensemble prior. For simplicity we assume a uniform prior over all the models, and we consider a wide array of different ensembles of these BNNs, including: ensembles of models with the same variable hyperparameter granularity; ensembles of models with the same number of nodes per layer; and ensembles of models with the same number of layers.
As detailed in \cref{f:all_mean_results}, from the predictions and the evidences associated with the ensembles we found: 1) The evidence--test set performance relations were very much the same as those found for the individual model results; 2) The best performance on the test set data overall was obtained with an ensemble of models: the combination of all models with layer granularity random variable hyperparameters achieved a test loss of 0.1732. The best performing individual model obtained a loss of 0.1791. 

\section{Conclusions}

\label{s:c}

In this paper we applied compromise-free Bayesian neural networks (BNNs) to the Boston housing dataset in order to explore the relationship between out-of-sample performance and the Bayesian evidence as first studied in \citet{mackay1992practical}, and how combining the predictions of different networks in a statistically sound way affects their performance. To numerically obtain samples of the full posterior and evidences associated with the training data and the networks, we used the nested sampling algorithm \PolyChord{} \citep{handley2015polychord1, handley2015polychord2}, to train models with up to $156$ trainable parameters. 
We considered a wide variety of models in our analysis: networks with either $\tanh$ or $\relu$ activation functions; networks with between zero and four hidden layers; and models trained using hierarchical Bayesian inference, i.e. models where the prior and likelihood standard deviations are modelled as random variables, as in \citet{mackay1992practical, neal2012bayesian, javid201921cm}. 

Our experiments demonstrate a proof-of-concept for two concrete principles: A) Using the Bayesian evidence one can quantify out-of-sample performance and generalisability from training data alone (and where this isn't the case, it is indicative that the model needs improvement); B) Ensembles of networks, which are obtained almost for free from previous runs can improve predictive performance.

A compromise-free approach is only ever intended as an initial step in a wider analysis. The purpose of solving the full numerical problem without approximation is twofold: 
First to see how far one can get with current computing resources, and hence to forecast how things may scale both now and in the future with more computing resources, time or money. 
Second, and more importantly, to use the results of the full solution as a foundation to an exciting area of model selection and ensembling.  We hope this paper will encourage the community to apply more approximate methods, and to consider larger, more practical applications, and test how the trends and characteristics presented here are affected by these regime changes. 
It is hoped that this work will be a springboard and inspiration for a profitable line of research into Bayesian neural networks in the context of model selection and model ensembling.

\vfill
\pagebreak
\section*{Broader Impact}

This work into compromise-free Bayesian neural networks is very much at a preliminary stage, but demonstrates that there are many benefits that a fully Bayesian numerical approach can bring to inference. 

If the further investigations into using the Bayesian evidence $\z$ to quantify out-of-sample generalisation from in-sample data show the observations of this paper to be robust, then the potential importance of this approach cannot be overstated. Whilst we do not advocate dispensing with out-of-sample testing data, there are many machine learning domains where obtaining such data is very hard or ethically impossible, such as autonomous vehicles or medical applications. In these settings the ability to select, deselect or marginalise over approaches that are unlikely to work well beyond the confines of a training dataset could prove essential, although relying blindly on the evidence could result in overfitting of the data.

Even in fields where testing and training data are plentiful, an orthogonal measure of generalisability could still prove invaluable to both improving performance, and reducing the effect of hidden biases in training data and systematic errors.

There is also scope for this work to further encourage the use of Bayesian modelling in machine learning in general. It is likely true that the observations regarding generalisability and ensembling obtained here are not neural network specific, and the ability to have principled uncertainty quantification (or certainty about one's uncertainty) is something that would be of great use in a wide variety of fields.

We do not necessarily advocate using a compromise-free approach for widespread industrial use in its present state. The computational cost is very resource-heavy (in both time, money, memory and energy). Whilst this can be brought down to human-scale training times using high-performance computing and the extensive parallelisation capability of \PolyChord{}, which may be sped up linearly to $n_\mathrm{CPUs}\sim\mathcal{O}(n_\mathrm{live})$, this is arguably not a very green approach. However in its current form it could be useful as an additional tool in the arsenal of a machine learning researcher for solving particularly stubborn supervised learning problems. 

As the technology behind the research is refined, the compromise-free approach may become more competitive in comparison with traditional training techniques. In such an instance, the products of this research will have a wide impact across a broad range of research and industrial applications.

\textbf{Cui bono:}  People wishing to gauge generalisation ability from training data; people concerned uncertainty quantification; people concerned with tying practical machine learning back to traditional statistics and theory; people concerned with approximations/assumptions used in training Bayesian models.

\textbf{Cui malo:} People concerned with the compute costs associated with training machine learning  models; people who are concerned primarily with optimal performance (e.g. lowest error) of ML models.

\begin{ack}
This work was performed using resources provided by the \href{http://www.csd3.cam.ac.uk/}{Cambridge Service for Data Driven Discovery (CSD3)} operated by the University of Cambridge Research Computing Service, provided by Dell EMC and Intel using Tier-2 funding from the Engineering and Physical Sciences Research Council (capital grant EP/P020259/1), and \href{www.dirac.ac.uk}{DiRAC funding from the Science and Technology Facilities Council}.
K.J.\ was supported by an STFC Impact Acceleration Account 2018 grant.
W.H.\ was supported by a Gonville \& Caius College Research fellowship and STFC IPS grant number G102229.
\end{ack}

\small
\setlength{\bibsep}{0.0pt}
\bibliographystyle{unsrtnat}
\bibliography{references}

\begin{thebibliography}{42}
\providecommand{\natexlab}[1]{#1}
\providecommand{\url}[1]{\texttt{#1}}
\expandafter\ifx\csname urlstyle\endcsname\relax
  \providecommand{\doi}[1]{doi: #1}\else
  \providecommand{\doi}{doi: \begingroup \urlstyle{rm}\Url}\fi

\bibitem[Krzywinski and Altman(2013)]{krzywinski2013points}
Martin Krzywinski and Naomi Altman.
\newblock Points of significance: Importance of being uncertain, 2013.

\bibitem[Ghahramani(2015)]{ghahramani2015probabilistic}
Zoubin Ghahramani.
\newblock Probabilistic machine learning and artificial intelligence.
\newblock \emph{Nature}, 521\penalty0 (7553):\penalty0 452, 2015.

\bibitem[Rumelhart et~al.(1985)Rumelhart, Hinton, and
  Williams]{rumelhart1985learning}
David~E Rumelhart, Geoffrey~E Hinton, and Ronald~J Williams.
\newblock Learning internal representations by error propagation.
\newblock Technical report, California Univ San Diego La Jolla Inst for
  Cognitive Science, 1985.

\bibitem[Denker et~al.(1987)Denker, Schwartz, Wittner, Solla, Howard, Jackel,
  and Hopfield]{denker1987large}
John Denker, Daniel Schwartz, Ben Wittner, Sara Solla, Richard Howard, Lawrence
  Jackel, and John Hopfield.
\newblock Large automatic learning, rule extraction, and generalization.
\newblock \emph{Complex systems}, 1\penalty0 (5):\penalty0 877--922, 1987.

\bibitem[Tishby et~al.(1989)Tishby, Levin, and Solla]{tishby1989consistent}
Naftali Tishby, Esther Levin, and Sara~A Solla.
\newblock Consistent inference of probabilities in layered networks:
  Predictions and generalization.
\newblock In \emph{International Joint Conference on Neural Networks},
  volume~2, pages 403--409, 1989.

\bibitem[Buntine and Weigend(1991)]{buntine1991bayesian}
Wray~L Buntine and Andreas~S Weigend.
\newblock Bayesian back-propagation.
\newblock \emph{Complex systems}, 5\penalty0 (6):\penalty0 603--643, 1991.

\bibitem[Denker and Lecun(1991)]{denker1991transforming}
John~S Denker and Yann Lecun.
\newblock Transforming neural-net output levels to probability distributions.
\newblock In \emph{Advances in neural information processing systems}, pages
  853--859, 1991.

\bibitem[MacKay et~al.(1994)]{mackay1994bayesian}
David~JC MacKay et~al.
\newblock Bayesian nonlinear modeling for the prediction competition.
\newblock \emph{ASHRAE transactions}, 100\penalty0 (2):\penalty0 1053--1062,
  1994.

\bibitem[MacKay(1992{\natexlab{a}})]{mackay1992bayesian}
David~JC MacKay.
\newblock Bayesian interpolation.
\newblock \emph{Neural computation}, 4\penalty0 (3):\penalty0 415--447,
  1992{\natexlab{a}}.

\bibitem[MacKay(1992{\natexlab{b}})]{mackay1992practical}
David~JC MacKay.
\newblock A practical bayesian framework for backpropagation networks.
\newblock \emph{Neural computation}, 4\penalty0 (3):\penalty0 448--472,
  1992{\natexlab{b}}.

\bibitem[MacKay(1992{\natexlab{c}})]{mackay1992evidence}
David~JC MacKay.
\newblock The evidence framework applied to classification networks.
\newblock \emph{Neural computation}, 4\penalty0 (5):\penalty0 720--736,
  1992{\natexlab{c}}.

\bibitem[Rasmussen(2003)]{rasmussen2003gaussian}
Carl~Edward Rasmussen.
\newblock Gaussian processes in machine learning.
\newblock In \emph{Summer School on Machine Learning}, pages 63--71. Springer,
  2003.

\bibitem[Neal(1992)]{neal1992bayesian}
Radford~M Neal.
\newblock Bayesian training of backpropagation networks by the hybrid monte
  carlo method.
\newblock Technical report, Citeseer, 1992.

\bibitem[Neal(1993)]{neal1993bayesian}
Radford~M Neal.
\newblock Bayesian learning via stochastic dynamics.
\newblock In \emph{Advances in neural information processing systems}, pages
  475--482, 1993.

\bibitem[de~Freitas et~al.(2000)de~Freitas, Niranjan, Gee, and
  Doucet]{freitas2000sequential}
JFG~de de~Freitas, Mahesan Niranjan, Andrew~H. Gee, and Arnaud Doucet.
\newblock Sequential monte carlo methods to train neural network models.
\newblock \emph{Neural computation}, 12\penalty0 (4):\penalty0 955--993, 2000.

\bibitem[de~Freitas(2003)]{de2003bayesian}
Jo{\~a}o Ferdinando~Gomes de~Freitas.
\newblock \emph{Bayesian methods for neural networks}.
\newblock PhD thesis, University of Cambridge, 2003.

\bibitem[Andrieu et~al.(2003)Andrieu, de~Freitas, Doucet, and
  Jordan]{andrieu2003introduction}
Christophe Andrieu, Nando de~Freitas, Arnaud Doucet, and Michael~I Jordan.
\newblock An introduction to mcmc for machine learning.
\newblock \emph{Machine learning}, 50\penalty0 (1-2):\penalty0 5--43, 2003.

\bibitem[Hinton and Van~Camp(1993)]{hinton1993keeping}
Geoffrey Hinton and Drew Van~Camp.
\newblock Keeping neural networks simple by minimizing the description length
  of the weights.
\newblock In \emph{in Proc. of the 6th Ann. ACM Conf. on Computational Learning
  Theory}. Citeseer, 1993.

\bibitem[Barber and Bishop(1998)]{barber1998ensemble}
David Barber and Christopher~M Bishop.
\newblock Ensemble learning in bayesian neural networks.
\newblock \emph{Nato ASI Series F Computer and Systems Sciences}, 168:\penalty0
  215--238, 1998.

\bibitem[Graves(2011)]{graves2011practical}
Alex Graves.
\newblock Practical variational inference for neural networks.
\newblock In \emph{Advances in neural information processing systems}, pages
  2348--2356, 2011.

\bibitem[Gal and Ghahramani(2016{\natexlab{a}})]{gal2016dropout}
Yarin Gal and Zoubin Ghahramani.
\newblock Dropout as a bayesian approximation: Representing model uncertainty
  in deep learning.
\newblock In \emph{international conference on machine learning}, pages
  1050--1059, 2016{\natexlab{a}}.

\bibitem[Gal and Ghahramani(2016{\natexlab{b}})]{gal2016theoretically}
Yarin Gal and Zoubin Ghahramani.
\newblock A theoretically grounded application of dropout in recurrent neural
  networks.
\newblock In \emph{Advances in neural information processing systems}, pages
  1019--1027, 2016{\natexlab{b}}.

\bibitem[Kendall and Gal(2017)]{kendall2017uncertainties}
Alex Kendall and Yarin Gal.
\newblock What uncertainties do we need in bayesian deep learning for computer
  vision?
\newblock In \emph{Advances in neural information processing systems}, pages
  5574--5584, 2017.

\bibitem[Higson et~al.(2018)Higson, Handley, Hobson, and
  Lasenby]{higson2018bayesian}
Edward Higson, Will Handley, Michael Hobson, and Anthony Lasenby.
\newblock Bayesian sparse reconstruction: a brute-force approach to
  astronomical imaging and machine learning.
\newblock \emph{Monthly Notices of the Royal Astronomical Society},
  483\penalty0 (4):\penalty0 4828--4846, 2018.

\bibitem[Handley et~al.(2015{\natexlab{a}})Handley, Hobson, and
  Lasenby]{handley2015polychord1}
WJ~Handley, MP~Hobson, and AN~Lasenby.
\newblock Polychord: nested sampling for cosmology.
\newblock \emph{Monthly Notices of the Royal Astronomical Society: Letters},
  450\penalty0 (1):\penalty0 L61--L65, 2015{\natexlab{a}}.

\bibitem[Handley et~al.(2015{\natexlab{b}})Handley, Hobson, and
  Lasenby]{handley2015polychord2}
WJ~Handley, MP~Hobson, and AN~Lasenby.
\newblock Polychord: next-generation nested sampling.
\newblock \emph{Monthly Notices of the Royal Astronomical Society},
  453\penalty0 (4):\penalty0 4384--4398, 2015{\natexlab{b}}.

\bibitem[Neal(2012)]{neal2012bayesian}
Radford~M Neal.
\newblock \emph{Bayesian learning for neural networks}, volume 118.
\newblock Springer Science \& Business Media, 2012.

\bibitem[Stone(1977)]{stone1977asymptotics}
Mervyn Stone.
\newblock Asymptotics for and against cross-validation.
\newblock \emph{Biometrika}, pages 29--35, 1977.

\bibitem[Efron(1979)]{efron1979computers}
Bradley Efron.
\newblock Computers and the theory of statistics: thinking the unthinkable.
\newblock \emph{SIAM review}, 21\penalty0 (4):\penalty0 460--480, 1979.

\bibitem[Li et~al.(1985)]{li1985stein}
Ker-Chau Li et~al.
\newblock From stein's unbiased risk estimates to the method of generalized
  cross validation.
\newblock \emph{The Annals of Statistics}, 13\penalty0 (4):\penalty0
  1352--1377, 1985.

\bibitem[Snoek et~al.(2012)Snoek, Larochelle, and Adams]{snoek2012practical}
Jasper Snoek, Hugo Larochelle, and Ryan~P Adams.
\newblock Practical bayesian optimization of machine learning algorithms.
\newblock In \emph{Advances in neural information processing systems}, pages
  2951--2959, 2012.

\bibitem[{Handley}(2019)]{2019JOSS....4.1414H}
Will {Handley}.
\newblock {anesthetic: nested sampling visualisation}.
\newblock \emph{The Journal of Open Source Software}, 4:\penalty0 1414, May
  2019.
\newblock \doi{10.21105/joss.01414}.

\bibitem[Goodfellow et~al.(2016)Goodfellow, Bengio, and
  Courville]{goodfellow2016deep}
Ian Goodfellow, Yoshua Bengio, and Aaron Courville.
\newblock \emph{Deep learning}.
\newblock MIT press, 2016.

\bibitem[Handley et~al.(2019)Handley, Lasenby, Peiris, and
  Hobson]{handley2019bayesian}
Will~J Handley, Anthony~N Lasenby, Hiranya~V Peiris, and Michael~P Hobson.
\newblock Bayesian inflationary reconstructions from planck 2018 data.
\newblock \emph{arXiv preprint arXiv:1908.00906}, 2019.

\bibitem[Buscicchio et~al.(2019)Buscicchio, Roebber, Goldstein, and
  Moore]{buscicchio2019label}
Riccardo Buscicchio, Elinore Roebber, Janna~M Goldstein, and Christopher~J
  Moore.
\newblock The label switching problem in bayesian analysis for gravitational
  wave astronomy.
\newblock \emph{arXiv preprint arXiv:1907.11631}, 2019.

\bibitem[Skilling(2006)]{skilling2006nested}
John Skilling.
\newblock Nested sampling for general bayesian computation.
\newblock \emph{Bayesian analysis}, 1\penalty0 (4):\penalty0 833--859, 2006.

\bibitem[Javid et~al.(2020)Javid, Handley, Hobson, and Lasenby]{javid201921cm}
Kamran Javid, Will Handley, Mike Hobson, and Anthony Lasenby.
\newblock Twentyoneflow: accelerated global 21cm signal emulation with
  tensorflow and bayesian neural networks.
\newblock \emph{MNRAS (In preparation, private communication)}, 2020.

\bibitem[{Handley} and {Millea}(2019)]{2019Entrp..21..272H}
Will {Handley} and Marius {Millea}.
\newblock {Maximum-Entropy Priors with Derived Parameters in a Specified
  Distribution}.
\newblock \emph{Entropy}, 21\penalty0 (3):\penalty0 272, Mar 2019.
\newblock \doi{10.3390/e21030272}.

\bibitem[Feroz et~al.(2009)Feroz, Hobson, and Bridges]{feroz2009multinest}
F~Feroz, MP~Hobson, and M~Bridges.
\newblock Multinest: an efficient and robust bayesian inference tool for
  cosmology and particle physics.
\newblock \emph{Monthly Notices of the Royal Astronomical Society},
  398\penalty0 (4):\penalty0 1601--1614, 2009.

\bibitem[Higson et~al.(2019)Higson, Handley, Hobson, and
  Lasenby]{higson2019nestcheck}
Edward Higson, Will Handley, Michael Hobson, and Anthony Lasenby.
\newblock nestcheck: diagnostic tests for nested sampling calculations.
\newblock \emph{Monthly Notices of the Royal Astronomical Society},
  483\penalty0 (2):\penalty0 2044--2056, 2019.

\bibitem[MacKay(2003)]{mackay2003information}
David~JC MacKay.
\newblock \emph{Information theory, inference and learning algorithms}.
\newblock Cambridge university press, 2003.

\bibitem[Kingma and Ba(2014)]{kingma2014adam}
Diederik~P Kingma and Jimmy Ba.
\newblock Adam: A method for stochastic optimization.
\newblock \emph{arXiv preprint arXiv:1412.6980}, 2014.

\end{thebibliography}

\appendix

\section{Additional detail for Bayesian framework}\label{s:further_bayes}
In this appendix we collect together additional material describing our Bayesian approach which we consider too much detail for the main paper, but may be of use to readers who are new to numerical Bayesian inference.

 \subsection{Propagating Bayesian errors from weights to predictions}\label{s:propagating}
 In \cref{s:bnn} we briefly commented that the posterior described by \cref{e:bayes} quantifies the error in our fitted network parameters $\theta$, and that this error may be ``forwarded onto a distribution on the predictions $y_\mathrm{pred}$''. More precisely, a predictive distribution for $y_\mathrm{pred}$, from unseen inputs $x_\mathrm{new}$ by the network is induced by the posterior, and may be computed by marginalisation~\citep{2019Entrp..21..272H}
\begin{align}
    P(y_\mathrm{pred}|x_\mathrm{new},D_\mathrm{train},\mathcal{M}) &= \int \delta(y_\mathrm{pred}-f(x_\mathrm{new};\theta)) P(\theta|D_\mathrm{train},\mathcal{M})\mathrm{d}{\theta}.
    \label{e:pred}\\
    &= \frac{\mathrm{d}}{\mathrm{d}y_\mathrm{pred}} \int\limits_{{f(x_\mathrm{new};\theta)}<y_\mathrm{pred}} P(\theta|D_\mathrm{train},\mathcal{M})\mathrm{d}{\theta}. \label{e:pred2}
\end{align}
If desired, this above distribution can be compressed into summary statistics such as a mean prediction and error bar
\begin{align}
    \widehat{y}_\mathrm{pred} &= \int P(\theta|D_\mathrm{train},\mathcal{M}) f(x_\mathrm{new};\theta), 
    \mathrm{d}{\theta}
    = \left\langle f(x_\mathrm{new};\theta)\right\rangle
    \label{e:mean}
    \\
    \sigma_{y, \mathrm{pred}}^2 &= \int P(\theta|D_\mathrm{train},\mathcal{M}) (f(x_\mathrm{new};\theta)-\widehat{y}_\mathrm{pred})^2 \mathrm{d}{\theta} = \mathrm{var}(f(x_\mathrm{new};\theta)).
    \label{e:var}
\end{align}                                                                                  

The above expressions \cref{e:pred,e:pred2,e:mean,e:var} are all conditioned on a specific network architecture model $\mathcal{M}$. 
One can of course use the Bayesian evidences to marginalise out this network dependence completely and obtain values for $\widehat{y}_\mathrm{pred}$ and $\sigma_{y, \mathrm{pred}}$ corresponding to the posterior distribution over network architectures. The equivalent of \cref{e:pred} becomes
\begin{equation}
    P(y|x,D_\mathrm{train}) = \sum_m P(y|x,D_\mathrm{train},m) P(m|D_\mathrm{train}),
    \label{e:pred_marg}
\end{equation}
with corresponding marginal summary statistics such as:
\begin{equation}
    \widehat{y}_\mathrm{pred} = \sum_m \widehat{y}_\mathrm{pred}(m) P(m|D_\mathrm{train}),
\end{equation}
where $\widehat{y}_\mathrm{pred}(m)$ are the corresponding means from \cref{e:mean} conditioned on model $m$. In using \cref{e:pred_marg}, one is effectively marginalising over an ensemble of networks, weighted by the quality of the fit as viewed by the data versus the model complexity.

\subsection{Bayesian inference in practice}\label{s:practice}

Whilst the likelihood in \cref{e:like} is Gaussian with respect to the data, it is highly non-Gaussian with respect to the parameters $\theta$. We must therefore use numerical Bayesian inference, for which the critical concept is that of sampling a distribution.

Sampling from a distribution provides a natural compression scheme, encoding the critical information in a general posterior $P(\theta)$ as a set of weighted samples drawn from it. From samples, one may perform the otherwise challenging but critical operations of marginalisation and transformation of distributions with ease. Marginalisation amounts to ignoring coordinates, and from a set of samples from $P(\theta)$ one can easily generate samples from an alternative distribution $P(q)$ where $q=q(\theta)$ by applying $q$ to each sample. This is very helpful for producing samples from a predictive distributions such as those given by \cref{e:pred,e:pred_marg}.

In the context of Bayesian neural networks, one can consider the traditional predictive procedure of using $y_\mathrm{pred}=f(x_\mathrm{new},\theta_*)$ as being extended to using a set of predictions given by $f(x_\mathrm{new},\theta)$ for the sampled values of $\theta$, each weighted by the corresponding $P(\theta)$ to give the posterior over $y_{\mathrm{pred}}$ (and with it summary statistics such as $\widehat{y}_\mathrm{pred}$ and $\sigma_{y, \mathrm{pred}}$).
One can think of marginalising over independently trained networks as extending this to using an ensemble of networks, and combining these estimates (weighted according to \cref{e:pred_marg}) to obtain $\widehat{y}_\mathrm{pred}$.
%moved here for arxiv version so it doesn't appear in middle of references
\begin{figure}
    \centerline{
        \includegraphics{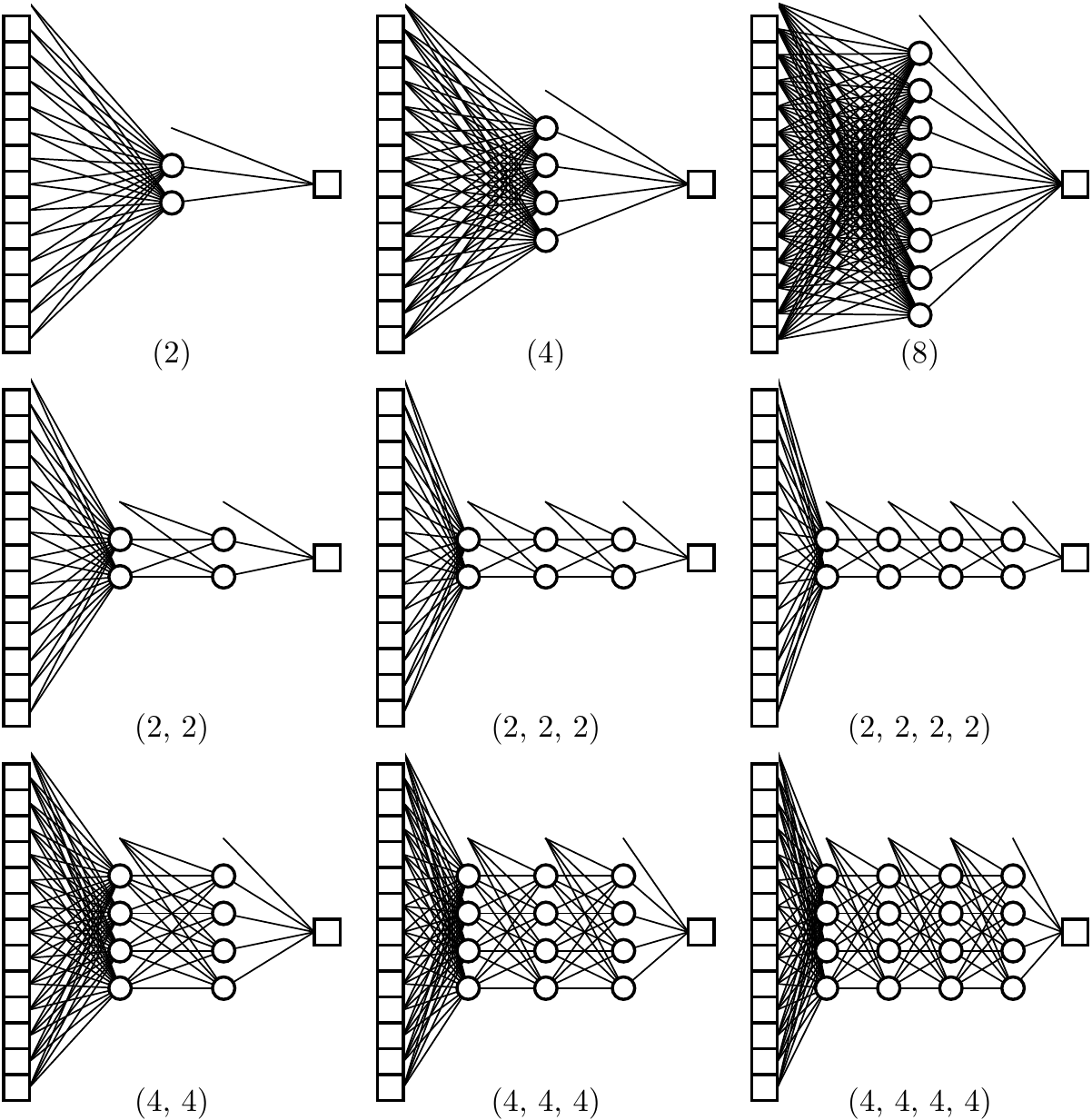}
    }
    \caption{The neural network architectures considered in this paper.
        These are graphical representations of \cref{e:nn} with inputs $x$ ($l_{0}=13$) and outputs $y$ ($l_{L}=1$) illustrated as squares on the left and right hand sides, intermediate node values $z$ as circles arranged in vertical layers $\ell$ and weights $w$ and biases $b$ as solid lines. Each architecture is summarised by a list of numbers in parentheses, giving the number of nodes in each hidden layer.
        \label{f:nn}
}
\end{figure}

\subsection{The prior}
\label{s:hbsh}
In order to begin the process of Bayesian inference, we must specify the prior. Throughout, we take the prior on the network parameters and bias terms to be independent normal distributions with zero mean and width $\sigma_i$ so each component $\theta_i$ has
\begin{equation}\label{e:nprior}
    P(\theta_{i}|\sigma_i,\mathcal{M}) = \frac{1}{\sqrt{2\pi}\sigma_i}\exp\left(-\frac{\theta_{i}^2}{2\sigma_i^2}\right).
\end{equation}
Note that in the traditional approach denoted by \cref{e:chi2} defining $R$ to be a function of the $l_2$ norm of $\theta$ was inspired from the use of a Gaussian prior over the network parameters. In order to fully specify the model, one must also specify the individual prior and likelihood widths $\sigma_i$ and $\sigma$, which play an analogous role to the regularisation parameters $\lambda_r$ and $\lambda_m$ in \cref{e:chi2} respectively. Since the data $x$ and $y$ can be (and are generally) whitened to have zero mean and unit variance (\cref{s:methods}), a not unreasonable choice is to set the likelihood variance to one. Setting $\sigma_i = 1 $ for all $i$ is equivalent to setting $\lambda_r=\frac{1}{2}$.

In the spirit of model comparison however, an extended strategy is to check whether treating $\sigma_i$ and $\sigma$ as free hyperparameters alongside the network parameters provides better performance and to see how the Bayesian evidence is affected. This necessitates specifying a further hyperprior on the widths, which we will in general take to be a gamma prior on the precision $\tau_i=\sigma_i^{-2}$
\begin{align}
    %P(\tau_i|\ga,\gb,\mathcal{M}) &= \frac{\gb^{\ga}}{\Gamma(\ga)} \tau_i^{\ga - 1} e^{-\gb \tau_i}\nonumber\\
    %\Rightarrow
    P(\sigma_i|\ga,\gb,\mathcal{M}) &= \frac{2\gb^{\ga}}{\Gamma(\ga)} \sigma_i^{-2\ga - 1} e^{-\gb/\sigma_i^2},\label{e:gnprior}
\end{align}
with an equivalent expression for a distribution on $\sigma$ with hyperparameters $\alpha$ and $\beta$.
This procedure of treating the widths $\sigma_i$ as additional parameters is common in hierarchical Bayesian inference, and is equivalent to letting the data decide the width of the distributions from which the network parameters are sampled. In the language of traditional neural network training, this is equivalent to letting the data decide the regularisation factor. For the likelihood function, treating the width $\sigma$ as a random variable essentially lets us estimate the data noise from the data themselves.

The downside to this hierarchical approach is that we still have to assign the values of the hyperparameters $\ga, \gb,\alpha,\beta$ for the hierarchical priors (discussed in \cref{s:methods}). An appropriately chosen hyperprior will be designed so that the choice of hyperparameters for the hierarchical prior has a diluted effect compared to choosing the hyperparameters for the base prior (i.e. $\sigma=\sigma_i=1$). We note that \citet{neal2012bayesian} used two-level hierarchical Bayesian inference to push this problem deeper: the hyperparameters of the hierarchical prior are themselves assigned a prior distribution, and it is this second-level hierarchical prior which has to have hyperparameters assigned deterministically. Ideally one would continue down the hierarchy until the Bayesian evidence tells us that we do not need to go any further, but in this work we only considered up to one-level hierarchical models, leaving a deeper analysis to future research.

\subsection{Weight space symmetry}
\label{s:wss}

For the hidden layers in a neural network, a degeneracy between the weights/biases in different nodes exists within a given layer \citep{higson2018bayesian}. In deep learning this is known as weight space symmetry \citep{goodfellow2016deep}. For a fully-connected feed-forward neural network, the degeneracy arises due to the fact that any node is just a linear combination of the outputs of the previous (usually followed by a non-linear activation). Thus no node within a layer is unique, and so is degenerate with all other nodes in that layer. This means that for a neural network with $\nl$ hidden layers, where the number of nodes in the layers is given by $(l_{1},...,l_{\nl})$, then the total degeneracy of the network is $\prod_{i=1}^{i=\nl}l_{i}!$. This degeneracy exponentially increases the size of parameter space to be explored without providing a better fit, and so should be avoided for computational efficiency whenever possible. 

This problem may be resolved by using a {\em forced identifiability prior\/} \citep{handley2019bayesian, buscicchio2019label}, which enforces an artificial ordering on degenerate parameters. When applied to the bias terms in a layer, this provides a labelling on the nodes and breaks the degeneracy between them.  

Usually when using a sampling algorithm which samples from the unit hypercube such as \PolyChord{}, one obtains a sequence of parameters ($\theta_{1},...,\theta_{\np}$) in physical space from their representations in the unit hypercube ($u_{1},...,u_{\np}$) using the inverse CDF function of the prior. When using a forced identifiability prior, an intermediate step enforces an ordering on the unit hypercube values, involving a reversed recurrence relation, starting by updating $u_{\np} \rightarrow u_{\np}^{1/\np}$ and then
$u_{i} \rightarrow u_{i}^{\frac{1}{i + 1}}u_{i+1}$. This enforces an ordering $u_{1} < u_{2} \ldots < u_{\np}$, thus breaking any switching degeneracy in $\theta_{i}$ values.

\subsection{Hierarchical priors}\label{s:hierarchical}

 For the hierarchical priors on the network parameter prior widths $\sigma_i$, we consider three granularities of prior:
\begin{description}
    \item[Single granularity] 
        One global precision hyperparameter $\sigma_1$ controlling the precision of all weights and biases in the network, with a Gamma prior on the precision with $\alpha_1=\beta_1=1$.
    \item[Layer granularity] 
        Two precision hyperparameters $\sigma_i$ per layer ($2(\nl+1)$ in total), one controlling the precision of the weights within a layer, the other of the bias parameters within a layer. A Gamma prior with $\ga = 1$, $\gb = 1$ is used for $i$ in the first hidden layer and for the bias nodes in each layer.  For subsequent hidden layers and the output layer, the weights of the $j^{\mathrm{th}}$ layer were assigned a Gamma distribution with $\ga = 1$, $\gb = 1 / l_{j-1}$, following the scaling arguments of \citet{neal2012bayesian} (see \cref{s:gran} for more detail).
    \item[Input size granularity] 
        All weights within a layer multiplying the same activation $z$ share a precision hyperparameter $\sigma_i$. Bias hyperparameters are shared per-layer as in layer granularity. The total number of variable prior hyperparameters for the network is therefore $\sum_{j = 0}^{j = \nl - 1} (l_{j}+1)$. The same scaling of the Gamma distributions as for layer granularity is adopted.
\end{description}
As an example of input size granularity consider the network with $(l_{1}, l_{2}) = (4, 4)$ (n.b. the input and output layers are $l_{0} = 13$ and $l_{3} = 1$ respectively). The three sets of biases for each layer each are assigned a Gamma hyperprior with $\ga = \gb = 1$. The weights in the first hidden layer have 13 Gamma hierarchical priors with these same hyperparameters. The second hidden layer has four Gamma hyperpriors with $\ga = 1$ and $\gb = 1 / l_{1} = 1 / 4$. The output node's weights have a separate hyperprior assigned to them all with $\ga = 1$ and $\gb = 1 / l_{1} = 1 / 4$. Thus the number of variable hyperparameters for this setup is $3 + 13 + 4 + 4 = 24$.  

\subsection{Implementing hierarchical Bayesian inference}\label{s:gran} 
\label{s:ihbi}

The single and layer granularities were used in \citet{mackay1992practical}, where he found that the former lead to a poor correlation between $\z$ and test set performance, when comparing models of different sizes. He argued that this was due to the fact that the input, outputs and hidden units had no reason to take the same scale of values, and thus scaling the weights associated with the different layers by the same factor (hyperparameter) was not the right thing to do. Thus he assigned one hyperparameter to the hidden unit weights, one to the hidden unit bias, and one for the output weights and biases (note this is slightly less granular than our implementation, which assigns separate hyperparameters for the output weights and bias). With this model he finds a much stronger correlation between $\z$ and test set performance, and an overall improvement in the models' performance. This is an example of the evidence not only being used as a proxy for out-of-sample performance, but also as an indicator that model performance can be improved in some way (though in general it does not give any indication of how to improve the model). The input size granularity takes inspiration from the automatic relevance determination (ARD) methodology introduced by \citet{mackay1994bayesian,neal2012bayesian}. The idea is to block out any inputs to a given layer which are not being used in learning the function mapping which the model represents. This is accomplished by inferring a large value for the precision (small value for the variance) associated with that input. Note that to the authors' knowledge, an in-depth analysis of the Bayesian evidence when considering granularities such as input size/ARD has not been done previously. 

\subsection{Gaussian processes as a prior over networks}
\label{s:gppn}

\citet{neal2012bayesian} introduces further insight into prior hyperparameters through his analysis of Gaussian processes and their relation to Bayesian neural networks. Neal finds that in the limit of an infinitely wide neural network, the network prior converges to a Gaussian process when the priors of the network weights are appropriately scaled. Neal shows that to prevent overfitting the data when building an arbitrarily large network, one must scale the variance of the weight priors according to the size of the previous layer (this argument does not apply to the first hidden layer, as the input layer nodes have no such restriction on their contribution to the subsequent layers). This also ensures the prior over functions has a finite variance. Thus in the models we consider with layer or input size variable prior hyperparameters, for layer $i+1$ we scale the scale parameters of all the Gamma (hierarchical) priors for the weights, by the size of the previous hidden layer $l_{i}$, i.e. $\gb \rightarrow \gb / l_{i} $.

\subsection{Explanation of nested sampling algorithm parameters}
For nested sampling, the number of live points $n_\mathrm{live}$ acts as a resolution parameter, with runtime typically scaling linearly with $n_\mathrm{live}$, and evidence and parameter estimation sampling error decreasing as the square root of the number of live points. Setting $n_\mathrm{live}=1,000$ gives a good balance between computational feasibility and evidence accuracy. For \PolyChord{}, the number of repeats $n_\mathrm{repeats}$ serves as a reliability parameter\footnote{In analogy with the inverse of \MultiNest{}'s efficiency \texttt{efr} \citep{feroz2009multinest}}; setting it too low is liable to generate algorithm-dependent biases, but there is a point beyond which setting it arbitrarily high brings no further gain~\cite{higson2019nestcheck}. 

\section{Detailed discussion of results}
\label{s:detailed_results}

\begin{table}
\centerline{%
\begin{tabular}{lccccc}
\toprule
names &  test loss &  test loss error &       $\log(\mathcal{Z})$ &  $\log(\mathcal{Z})$ error & dimensionality \\
\midrule
br               &   0.3415 &         0.0070 & -294.16 &     0.11 & 14 \\
sh sv         &   0.3416 &         0.0036 & -201.16 &     0.14 & 16 \\
lh sv          &   0.3418 &         0.0036 & -202.12 &     0.14 & 17 \\
ih sv        &   0.3416 &         0.0036 & -209.32 &     0.15 & 28 \\
(2)             &   0.2746 &         0.0088 & -286.75 &     0.12 & 31 \\
r (2)           &   0.2526 &         0.0080 & -278.73 &     0.13 & 31\\
sh sv (2)       &   0.2181 &         0.0059 & -149.09 &     0.70 & 33 \\
lh sv (2)       &   0.2090 &         0.0049 & -141.35 &     0.33 & 36 \\
ih sv (2)       &   0.2214 &         0.0045 & -150.62 &     0.39 & 49 \\
(4)             &   0.2711 &         0.0090 & -287.61 &     0.11 & 61 \\
r (4)           &   0.2336 &         0.0085 & -281.28 &     0.13 & 61 \\
sh sv (4)       &   0.1999 &         0.0074 & -135.05 &     0.75 & 63 \\
lh sv (4)       &   0.1982 &         0.0086 & -121.86 &     1.06 & 66 \\
ih sv (4)       &   0.2021 &         0.0073 & -124.91 &     1.29 & 81 \\
(8)             &   0.2688 &         0.0092 & -291.26 &     0.11 & 121 \\
r (8)           &   0.2216 &         0.0086 & -287.74 &     0.14 & 121 \\
sh sv (8)       &   0.2082 &         0.0068 & -127.67 &     2.57 & 123 \\
lh sv (8)       &   0.1864 &         0.0069 & -108.84 &     1.16 & 126 \\
ih sv (8)       &   0.1913 &         0.0066 & -104.99 &     1.71 & 145 \\
(2, 2)           &   0.2773 &         0.0095 & -289.02 &     0.13 & 37 \\
r (2, 2)         &   0.2509 &         0.0085 & -279.56 &     0.12 & 37 \\
sh sv (2, 2)     &   0.1983 &         0.0072 & -138.37 &     0.70 & 39 \\
lh sv (2, 2)     &   0.2042 &         0.0063 & -130.96 &     0.48 & 44 \\
ih sv (2, 2)     &   0.2152 &         0.0067 & -139.92 &     0.75 & 58 \\
(4, 4)           &   0.2722 &         0.0101 & -289.18 &     0.10 & 81 \\
r (4, 4)         &   0.2311 &         0.0098 & -281.92 &     0.13 & 81 \\
sh sv (4, 4)     &   0.1877 &         0.0084 & -116.07 &     1.23 & 83 \\
lh sv (4, 4)     &   0.1791 &         0.0081 & -115.37 &     1.74 & 88 \\
ih sv (4, 4)     &   0.2033 &         0.0078 & -115.78 &     1.38 & 106 \\
(2, 2, 2)         &   0.2779 &         0.0096 & -290.47 &     0.13 & 43 \\
r (2, 2, 2)       &   0.2558 &         0.0086 & -280.53 &     0.15 & 43 \\
sh sv (2, 2, 2)   &   0.2032 &         0.0077 & -137.25 &     0.76 & 45 \\
lh sv (2, 2, 2)   &   0.1995 &         0.0056 & -129.39 &     0.57 & 52 \\
ih sv (2, 2, 2)   &   0.2078 &         0.0061 & -141.26 &     0.92 & 67 \\
(4, 4, 4)         &   0.2730 &         0.0104 & -289.93 &     0.13 & 101 \\
r (4, 4, 4)       &   0.2335 &         0.0100 & -282.17 &     0.13 & 101 \\
sh sv (4, 4, 4)   &   0.2108 &         0.0081 & -116.89 &     1.57 & 103 \\
lh sv (4, 4, 4)   &   0.1801 &         0.0068 & -116.27 &     2.08 & 110 \\
ih sv (4, 4, 4)   &   0.1923 &         0.0080 & -117.12 &     1.01 & 131 \\
(2, 2, 2, 2)       &   0.2749 &         0.0094 & -291.49 &     0.13 & 49 \\
r (2, 2, 2, 2)     &   0.2589 &         0.0084 & -281.74 &     0.13 & 49 \\
sh sv (2, 2, 2, 2) &   0.2056 &         0.0068 & -137.83 &     0.91 & 51 \\
lh sv (2, 2, 2, 2) &   0.1963 &         0.0057 & -132.31 &     0.47 & 60 \\
ih sv (2, 2, 2, 2) &   0.2103 &         0.0068 & -141.68 &     0.73 & 76 \\
(4, 4, 4, 4)       &   0.2724 &         0.0105 & -290.34 &     0.21 & 121 \\
r (4, 4, 4, 4)     &   0.2387 &         0.0098 & -282.29 &     0.14 & 121 \\
sh sv (4, 4, 4, 4) &   0.2158 &         0.0077 & -118.75 &     1.11 & 123 \\
lh sv (4, 4, 4, 4) &   0.2035 &         0.0074 & -120.24 &     2.02 & 132 \\
ih sv (4, 4, 4, 4) &   0.2197 &         0.0073 & -125.18 &     1.93 & 156 \\
\bottomrule
\end{tabular}
}
\caption{Test loss and evidence comparison for all the individual BNNs. The losses and evidences ($\mathcal{Z}$) are averages obtained over the ten different data randomisations. The quoted errors are the standard deviations of the mean estimates. The key for the model names is as follows: r denotes $\relu$ activation functions were used in the hidden layers of the network ($\tanh$ otherwise). sh denotes a single random variable prior hyperparameter model, lh denotes layer granularity, while ih represents input size granularity. sv means the likelihood hyperparameter (variance) was also treated as variable. br (i.e. the first row) denotes Bayesian linear regression i.e. no hidden layers and fixed hyperparameters. The numbers in parentheses denote the number of nodes per hidden layer.
\label{t:non_combined_averages}}
\end{table}

In this appendix we present a more detailed breakdown of the results described in \cref{s:results}.

 Throughout our analysis we focussed on the average values obtained over the ten instances of each BNN as mentioned in \cref{s:methods}, but note that when we inspected the results from single instances of the randomisations, the same trends appeared albeit less coherently. \cref{t:non_combined_averages}  gives a summary of the average results obtained for the 49 different models, giving the test set loss values (mean squared error), the BNN's estimate of the error of the test losses, propagated through the error on the model predictions, $\sigy$, as well as the log evidences and their errors obtained from inferences. The dimensionalities include any variable hyperparameters involved in the analyses. Note that the four models with no hidden layers points consistently have test losses around 0.35, much higher than the other models considered. This emphasises the importance of deep networks, even when variable hyperparameters are used. For the rest of the analyses we do not consider these results obtained from the no hidden layer networks. 

\cref{f:non_comb_mean_results} shows the evidence versus the test loss for the 45 BNNs with hidden layers, from which one sees a clear separation between the BNNs with no variable hyperparameter (with either $\tanh$ or $\relu$ activations) and the variable hyperparameter BNNs (with $\tanh$ activations). Thus, straight away it is clear that the added complexity associated with variable hyperparameters is captured in the evidence values, but also provides an increase in performance.  
 
Much of the improvement in evidence value can be understood by the fact that when hyperparameters are not varied the likelihood variance $\sigma$ is set to unity, when in fact as shown in \cref{f:posteriors} its desired value of $\sigma\sim3.4\pm0.2$ is statistically significantly larger than this.

We now focus on different cross-sections of the set of models considered to get more insight into the evidence--test set loss relation for different subsets of the models.

\subsection{$\tanh$ and $\relu$ models}
\label{s:tanhreluresults}
\begin{figure}
  \begin{center}
      \includegraphics{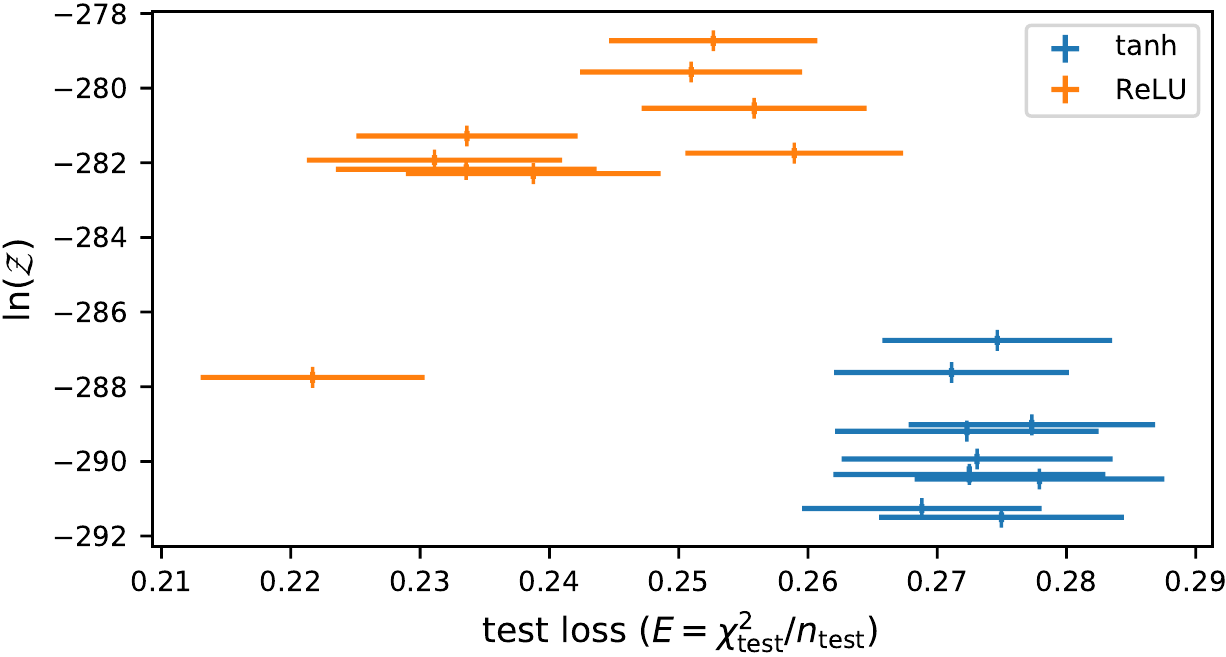}
  \caption{Detail of \cref{f:non_comb_mean_results} focussing on $\tanh$ (blue points) or $\relu$ (orange points) activation functions, and no variable hyperparameters.}
\label{f:non_comb_tanh_relu_mean_results}
  \end{center}
\end{figure}

We first focus on the models with no variable hyperparameters in \cref{f:non_comb_mean_results}. \cref{f:non_comb_tanh_relu_mean_results} shows that the models with $\relu$ activations consistently outperform the $\tanh$ models, and almost always provide a higher value of $\z$, as mentioned in the main text.
% This is quite a surprising result since {\em a-priori\/}, one may expect the opposite to occur, since $\tanh$ is a more non-linear function, one may expect it to perform well for the small networks considered here. In traditional neural network training, two of the key reasons why $\relu$ is a popular choice are that: 1) its derivative is fast and easy to calculate which is crucial for backward propagation training and 2) the fact that non-positive weights are shut out (their value and derivative are both zero) means that a side-effect of using $\relu$ is that it provides a form of regularisation, which can be helpful in large networks. Neither of these considerations are applicable to the BNNs we consider here, however, since \PolyChord{} does not use derivative information, and only small networks are considered. The only obvious potential benefit of using $\relu$ for these BNNs is that the function does not saturate for input values large in magnitude as $\tanh$ does. Regardless of why $\relu$ so consistently outperforms the $\tanh$ models, the evidence clearly picks up on its superiority. 

\begin{figure}
  \begin{center}
    \includegraphics{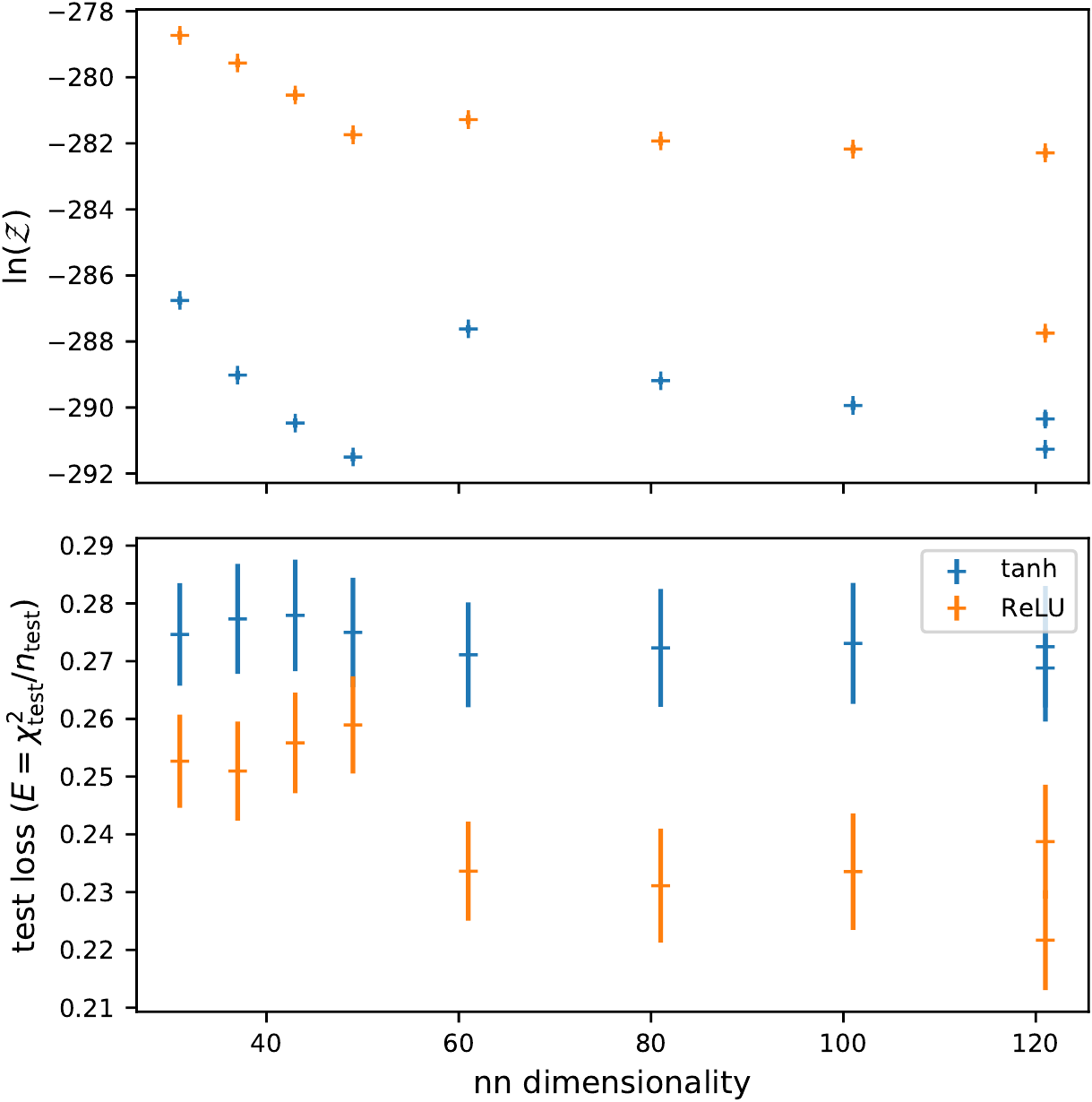}  
    \caption{Log Bayesian evidence (top plot) and test loss values (bottom plot) versus NN model dimensionality, for all individual BNNs with $\tanh$ (blue points) or $\relu$ (orange points) activation functions, with fixed hyperparameters. Note that, as in \cref{f:non_comb_sh_lh_ih_sv_mean_results_mean_Z_vs_dims_mean_loss_vs_dims}, model dimensionality is the total number of model parameters, including network and prior hyperparameters.
      }
\label{f:non_comb_tanh_relu_mean_Z_vs_dims_mean_loss_vs_dims}
  \end{center}
\end{figure}

\cref{f:non_comb_tanh_relu_mean_Z_vs_dims_mean_loss_vs_dims} shows the evidence versus the BNN dimensionality (top plot) and test loss versus the BNN dimensionality (bottom plot), for the fixed hyperparameter BNNs. The evidence seems to behave similarly (minus an offset) for both activations as a function of the model dimensionality.
Looking at the results for the two different activations separately, referring back to \cref{f:non_comb_tanh_relu_mean_results} there is no trend between evidence and test set performance for the different sized networks. From \cref{f:non_comb_tanh_relu_mean_Z_vs_dims_mean_loss_vs_dims} it appears that the $\relu$ models get better as they grow in size, but no such trend exists for the $\tanh$ models.

\subsection{Variable hyperparameter models}
\label{s:stochasticresults}
We now focus on the variable hyperparameter models in \cref{f:non_comb_mean_results}.
There is a marked increase in performance (i.e. decrease in test loss) in comparison with the fixed hyperparameter models, and a corresponding increase in Bayesian evidence. As can be seen in \cref{f:posteriors}, this is predominantly driven by the fact that the likelihood variance $\sigma$ preferred by the data is significantly different from the value chosen (unity) when the hyperparameters are fixed.

\begin{figure}
  \begin{center}
  \includegraphics{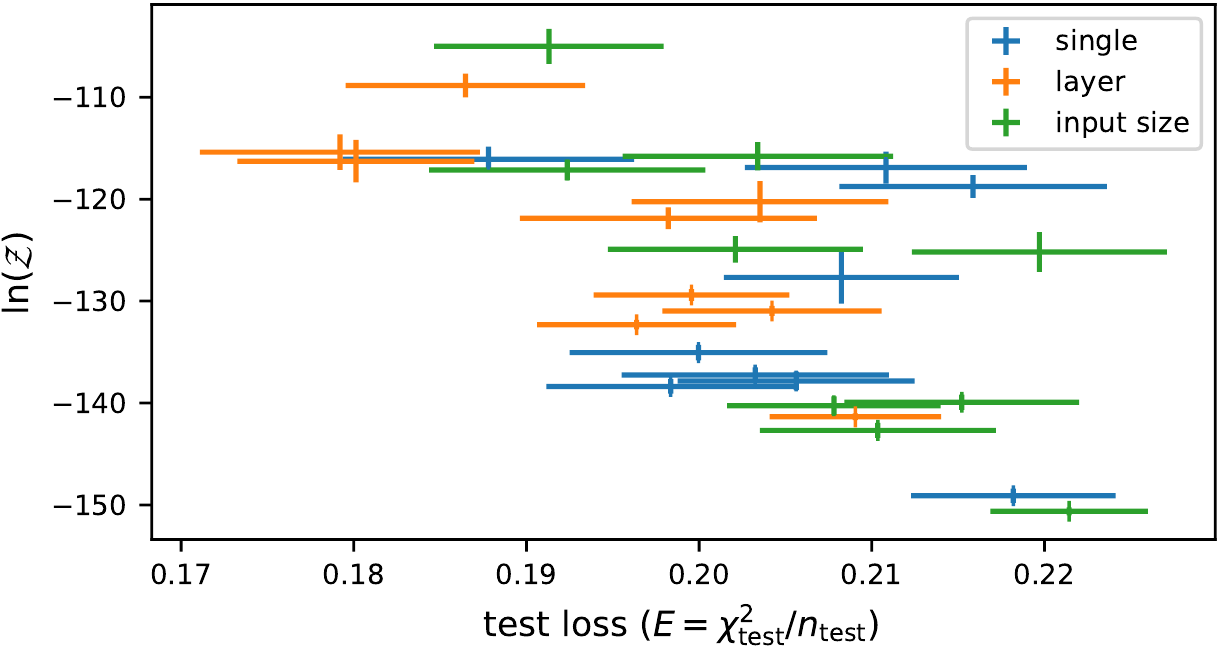}
  \caption{Detail of \cref{f:non_comb_mean_results} focussing on the cases with variable hyperparameters with single (blue points), layer (orange points), or input size granularity.}
\label{f:non_comb_sh_lh_ih_sv_mean_results}
  \end{center}
\end{figure}

\cref{f:non_comb_sh_lh_ih_sv_mean_results} shows the evidence versus test loss, and \cref{f:non_comb_sh_lh_ih_sv_mean_results_mean_Z_vs_dims_mean_loss_vs_dims} shows the evidence versus the BNN dimensionality (top plot), and test loss versus the BNN dimensionality (bottom plot), for models with different granularities of variable prior hyperparameters.
Looking at all the variable hyperparameter models as one (i.e. ignoring the colour-coding in 
\cref{f:non_comb_sh_lh_ih_sv_mean_results_mean_Z_vs_dims_mean_loss_vs_dims}), the correlation and symmetries remain, but the peaks/troughs are much less apparent in comparison with the same patterns within a colour class.

\subsubsection{Single random variable hyperparameter runs}
\label{s:shr}

Focusing first on the single random variable hyperparameter models, we see some correlation between Bayesian evidence and test set performance. We also see a remarkable symmetry between the log evidence as a function of NN dimensionality and the test set performance as a function of the same parameter. Furthermore, there appears to be a peak in the Bayesian evidence, which is fast to increase (with the BNN dimension), but relatively slow in decline. This is a well-known trend in model selection, called a Bayesian cliff, and was observed in \citet{mackay1992practical}. The corresponding dip in test set performance is less well-pronounced, but still arguably there and could be interpreted as a plateau. 

\subsubsection{Layer granularity runs}
\label{s:lhr}

For the layer random variable hyperparameter models, more of a correlation between test set performance and Bayesian evidence is present. Similar to the single random variable hyperparameter case, the $\z$--BNN dimensionality and test set performance--BNN dimensionality symmetry also appears. The corresponding peaks and troughs are also there, but are less clear-cut (more plateau-like). For layer granularity models the test set performance was better than the equivalent models with single granularity in most cases. The evidence values were also higher for the former in general, indicating that it correctly captured the superior performance. 

\subsubsection{Input size random variable hyperparameter runs}
\label{s:ihr}

Surprisingly, the input size random variable hyperparameter models consistently underperform the layer models (of the same model dimensionality). Since this model is more granular than the layer models, one may expect it to perform at least as well. Thus, one can only attribute this to the (more) complex parameter space not being explored as well, or, the model overfitting to the training data. Nevertheless, the input size results show a good correlation between test performance and evidence, a strong symmetry in evidence and test performance when plotted against BNN dimensionality, and arguably, the most well-formed peaks/troughs in the corresponding Figures.

\begin{table}
\centerline{%
\begin{tabular}{lcccccccc}
\toprule
  names &  $E_{\mathrm{tr, l}}$ &  $E_{\mathrm{tr, i}}$ &  $n_{\mathrm{l}}$ &  $n_{\mathrm{i}}$ &  $\widehat{\sigma_{\mathrm{p,l}}}$ &  $\widehat{\sigma_{\mathrm{p,l}}}$ error &  $\widehat{\sigma_{\mathrm{p,i}}}$ &  $\widehat{\sigma_{\mathrm{p,i}}}$ error \\
\midrule
          (2) &                0.1120 &                0.1270 &               4 &              17 &                              2.696 &                                    1.967 &                              1.650 &                                    1.991 \\
          (4) &                0.0598 &                0.0593 &               4 &              19 &                              4.763 &                                    1.036 &                              2.682 &                                    0.432 \\
          (8) &                0.0326 &                0.0312 &               4 &              23 &                              4.698 &                                    0.740 &                              3.314 &                                    0.758 \\
      (2, 2) &                0.0926 &                0.0901 &               6 &              20 &                              2.793 &                                    2.818 &                              1.829 &                                    0.583 \\
      (4, 4) &                0.0459 &                0.0502 &               6 &              24 &                              2.529 &                                    0.677 &                              1.846 &                                    0.495 \\
    (2, 2, 2) &                0.0946 &                0.0911 &               8 &              23 &                              3.103 &                                    2.366 &                              1.836 &                                    0.507 \\
    (4, 4, 4) &                0.0476 &                0.0512 &               8 &              29 &                              2.465 &                                    0.541 &                              2.566 &                                    1.262 \\
 (2, 2, 2, 2) &                0.0924 &                0.0895 &              10 &              26 &                              3.470 &                                    1.947 &                              1.910 &                                    0.623 \\
 (4, 4, 4, 4) &                0.0493 &                0.0525 &              10 &              34 &                              5.551 &                                    2.113 &                              1.787 &                                    0.318 \\
\bottomrule
\end{tabular}
}
\caption{Comparison of layer and input size granularity random variable prior hyperparameter models on training data. The training losses ($E_{\mathrm{tr,} \cdot}$) are averages obtained over the ten different data randomisations. $\sigma_{\mathrm{p}}$ denotes the standard deviation of the prior (i.e. the variable prior hyperparameters), and hats denote their average values. The quoted errors are the standard deviations of the mean estimates. The numbers in parentheses in the left-hand column specify the number of nodes in each hidden layer of the network, while $n_{\mathrm{l}}$ and $n_{\mathrm{i}}$ denote the number of variable hyperparameters in layer and input size models respectively.}
\label{t:lh_ih_tr}
\end{table}

Further investigation into the layer and input size granularity performances is warranted, since the latter has been shown to do better in the past \citep{neal2012bayesian} and other recent works \citep{javid201921cm}. We first checked the training set performance, which was similar between the two granularities in terms of mean squared errors (see \cref{t:lh_ih_tr}). Next we looked at the values of the prior standard deviations ($ \propto 1 / \sqrt{\lambda_{r}}$ where $\lambda_{r}$ is the regularisation constant in the traditional terminology). In all but one case, the input size standard deviations were smaller than those obtained from the layer granularity models, suggesting the former is a more regularised model. This is perhaps surprising, as the underperformance on the test data of the input size models would suggest the training data is being overfit. The most plausible explanation therefore is that the input size models are on average shutting off nodes important in making accurate predictions on the test data, and more so than the layer granularity models, while doing a similarly good job on the training data. 

As mentioned previously, \citet{mackay1992practical} found that when a correlation between $\z$ and test performance was not present, then the model could be improved in some way, and that the improved model exhibited this correlation more. One could argue the same trend has appeared in this work. In the previous section, we considered models with no variable hyperparameters, which for a given activation, showed no trend between $\z$ and test performance for different sized models. However once variable hyperparameters were included, correlations appeared, and so did an overall increase in the performance of the models.

\subsubsection{Analysis of results by model size}

\begin{figure}
  \begin{center}
  \includegraphics{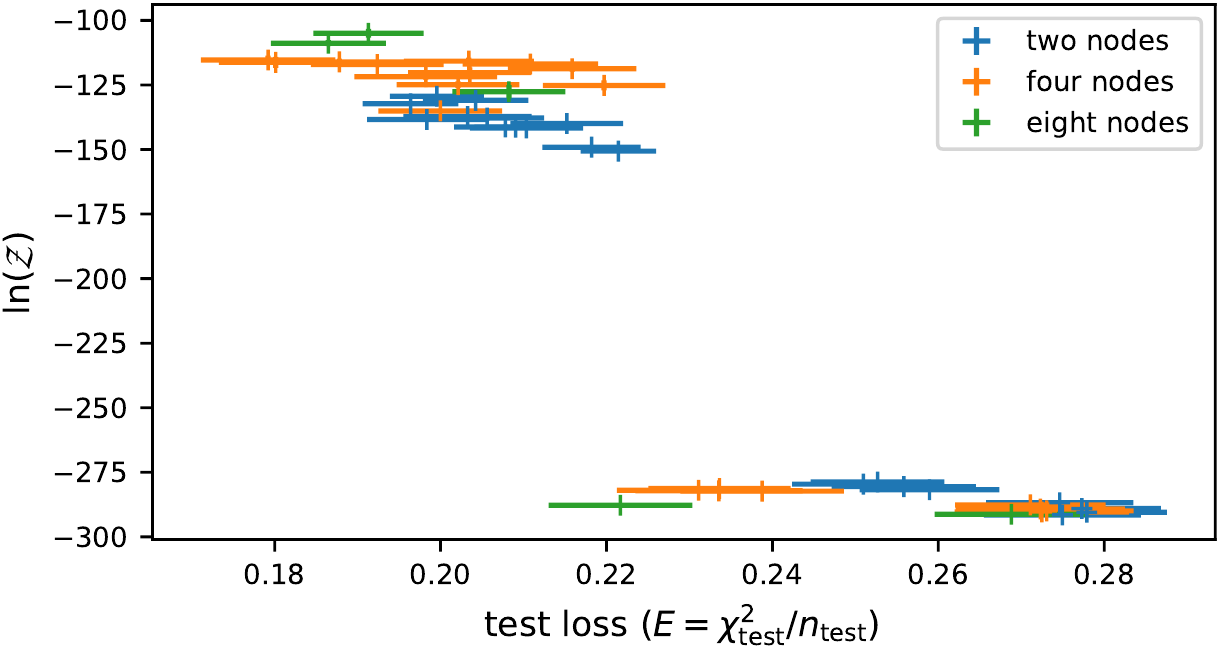}
  \caption{Log Bayesian evidence versus test loss for all individual BNNs with two node-wide layers (blue points), four node-wide layers (orange points), or eight node-wide layers (green points).}
\label{f:non_comb_2n_4n_8n_mean_results}
  \end{center}
\end{figure}

\begin{figure}
  \begin{center}
    \includegraphics{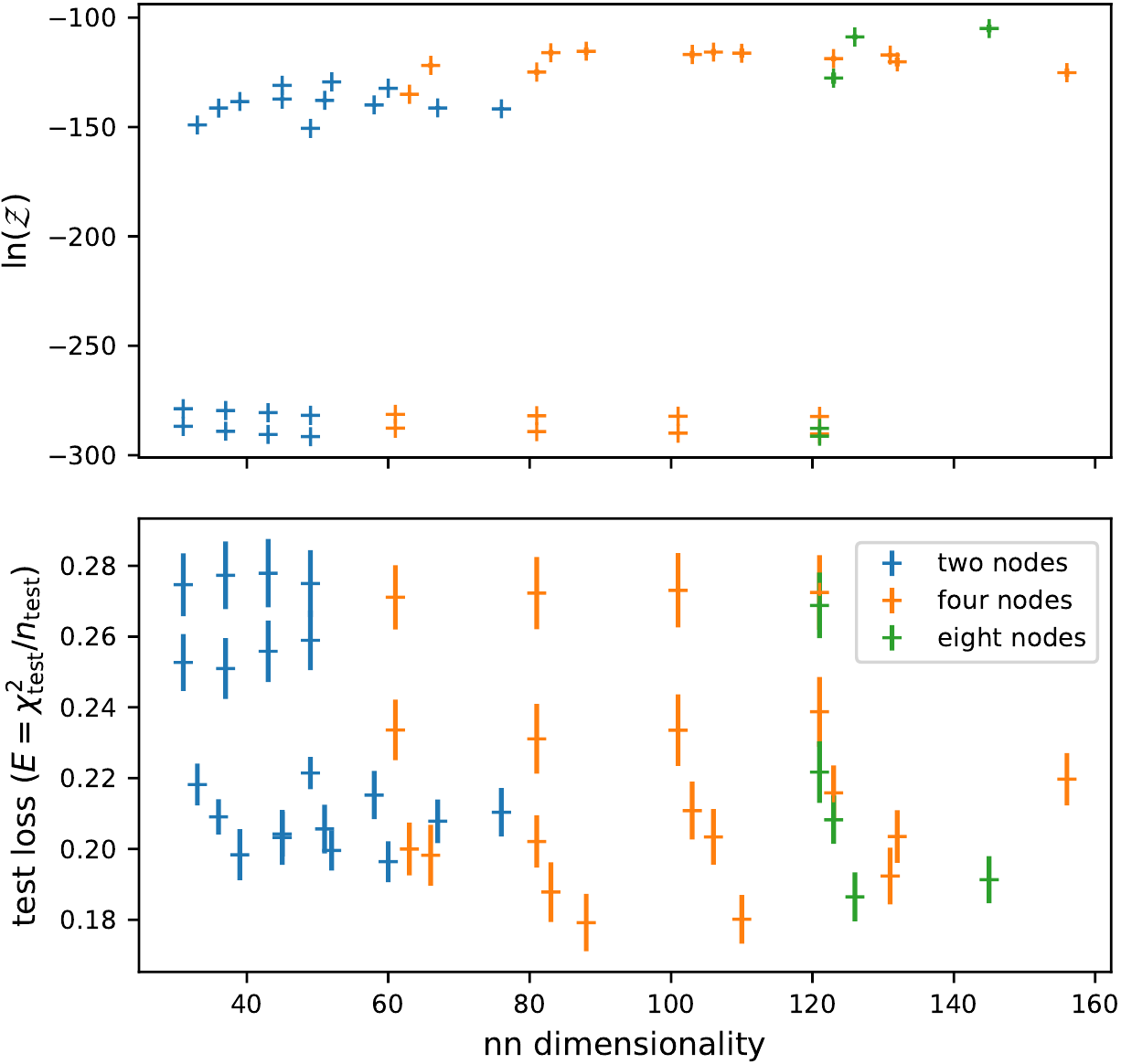}
  \caption{Log Bayesian evidence versus NN dimensionality (top plot) and test loss versus NN dimensionality (bottom plot), for all individual BNNs with two node-wide layers (blue points), four node-wide layers (orange points), or eight node-wide layers (green points).}
\label{f:non_comb_2n_4n_8n_mean_results_mean_Z_vs_dims_mean_loss_vs_dims}
  \end{center}
\end{figure}

Looking at different cross sections of the set of models by aggregating the models in terms of their size, we first look at all the $n$-node architectures, i.e. all the models which have $n$-nodes in their hidden layers. For the two-node models, \cref{f:non_comb_2n_4n_8n_mean_results} shows that the correlation between $\z$ and test performance is present for this subset, with clear modes corresponding to $\tanh$ models with fixed hyperparameters, $\relu$ models, and the variable hyperparameter models. The top and bottom plots of \cref{f:non_comb_2n_4n_8n_mean_results_mean_Z_vs_dims_mean_loss_vs_dims} 
shows that $\z$ and test set performance symmetries are also present, but the peaks/troughs/plateaus are more convoluted than the ones seen in \cref{s:stochasticresults}. The four and eight-node architectures show similar results.
When considering models grouped together by how many hidden layers they contain (one, two, three or four), the same overall patterns seem to exist, but there seems to be mainly two modes of separation: fixed hyperparameter versus variable hyperparameter models, as was the case in \cref{f:non_comb_mean_results}. For one layer, all but the one-layer model with eight nodes show prominent peaks/troughs associated with Occam's hill and test set performance versus BNN dimensionality. The same can be said for the 2, 3, and 4 layer models, but said relations are less clear-cut.

\subsection{Results for ensembled models}
\label{s:combinedresults}

\begin{table}
\centerline{%
\begin{tabular}{lcccc}
\toprule
names &  test loss &  test loss error &       $\log(\mathcal{Z})$ &  $\log(\mathcal{Z})$ error \\
\midrule
1l         &   0.2731 &         0.0089 & -287.49 &     0.09 \\
2l         &   0.2745 &         0.0098 & -289.09 &     0.08 \\
3l       &   0.2742 &         0.0101 & -290.16 &     0.10 \\
4l        &   0.2718 &         0.0101 & -290.76 &     0.16 \\
2n         &   0.2739 &         0.0091 & -288.01 &     0.10 \\
4n        &   0.2706 &         0.0096 & -288.68 &     0.08 \\
an         &   0.2724 &         0.0093 & -288.40 &     0.07 \\
1l r       &   0.2491 &         0.0081 & -279.75 &     0.12 \\
2l r       &   0.2459 &         0.0089 & -280.17 &     0.11 \\
3l r     &   0.2452 &         0.0093 & -281.05 &     0.13 \\
4l r      &   0.2479 &         0.0092 & -281.97 &     0.10 \\
2n r       &   0.2511 &         0.0083 & -279.61 &     0.09 \\
4n r      &   0.2325 &         0.0094 & -281.83 &     0.07 \\
an r       &   0.2463 &         0.0086 & -280.32 &     0.08 \\
1l aa      &   0.2491 &         0.0081 & -280.44 &     0.12 \\
2l aa      &   0.2459 &         0.0089 & -280.86 &     0.11 \\
3l aa    &   0.2452 &         0.0093 & -281.74 &     0.13 \\
4l aa     &   0.2479 &         0.0092 & -282.66 &     0.10 \\
2n aa      &   0.2523 &         0.0082 & -280.61 &     0.11 \\
4n aa     &   0.2326 &         0.0094 & -282.52 &     0.07 \\
an aa      &   0.2456 &         0.0086 & -281.28 &     0.09 \\
1l sh sv   &   0.2104 &         0.0072 & -128.76 &     2.57 \\
2l sh sv   &   0.1877 &         0.0084 & -116.77 &     1.23 \\
3l sh sv &   0.2108 &         0.0081 & -117.58 &     1.57 \\
4l sh sv  &   0.2158 &         0.0077 & -119.44 &     1.11 \\
2n sh sv   &   0.1985 &         0.0074 & -137.83 &     0.55 \\
4n sh sv  &   0.2140 &         0.0076 & -116.81 &     1.16 \\
an sh sv   &   0.2116 &         0.0075 & -117.62 &     1.16 \\
1l lh sv   &   0.1862 &         0.0070 & -109.94 &     1.16 \\
2l lh sv   &   0.1791 &         0.0081 & -116.07 &     1.74 \\
3l lh sv &   0.1801 &         0.0068 & -116.96 &     2.08 \\
4l lh sv  &   0.2035 &         0.0074 & -120.93 &     2.02 \\
2n lh sv   &   0.2228 &         0.0111 & -130.51 &     0.47 \\
4n lh sv  &   0.1797 &         0.0069 & -116.19 &     1.75 \\
an lh sv   &   0.1732 &         0.0075 & -111.03 &     1.16 \\
1l ih sv   &   0.1913 &         0.0066 & -106.09 &     1.71 \\
2l ih sv   &   0.2033 &         0.0078 & -116.47 &     1.38 \\
3l ih sv &   0.1923 &         0.0080 & -117.82 &     1.01 \\
4l ih sv  &   0.2197 &         0.0073 & -125.87 &     1.93 \\
2n ih sv   &   0.2074 &         0.0073 & -140.81 &     0.59 \\
4n ih sv  &   0.2052 &         0.0070 & -116.88 &     1.28 \\
an ih sv   &   0.2012 &         0.0064 & -107.19 &     1.71 \\
\bottomrule
\end{tabular}
}

\caption{Test loss and evidence comparisons for all the combined BNNs. The key for the model names is as follows (including the notation in \cref{t:non_combined_averages}: $X$l denotes the combination of all networks with $X$ layers. $X$n models combine all networks with $X$ nodes in the hidden layers. Note an denotes all number of nodes i.e. all model sizes combined, and aa denotes all activation functions, i.e. combining models with either $\tanh$ or $\relu$ activations in their hidden layers.
\label{t:combined_averages}}
\end{table}

As mentioned in \cref{s:practice}, the posterior distributions of different models can be combined (ensembled) such that one gets a posterior corresponding to a model in which one considers all of these models at once when training the data. We assign a uniform prior over models for all combined analyses, showing no preference for any particular model {\em a-priori\/}. We obtain the evidences and predictions associated with these combined models and analyse the $\z - \e$ relation for these ensembles, and compare the quality of their predictions relative to the individual models. We consider many different combinations of models, across various cross sections of the wide array of models used in the individual analyses. \cref{t:combined_averages} gives a full breakdown of the different ensembles considered, but to summarise, the combinations of models we consider broadly cover the following cross sections (and their supersets): different sized models with either the same or different activation functions; models with the same variable hyperparameter granularity; models with the same number of nodes per layer; and models with the same number of layers.

\begin{figure}
  \begin{center}
  \includegraphics{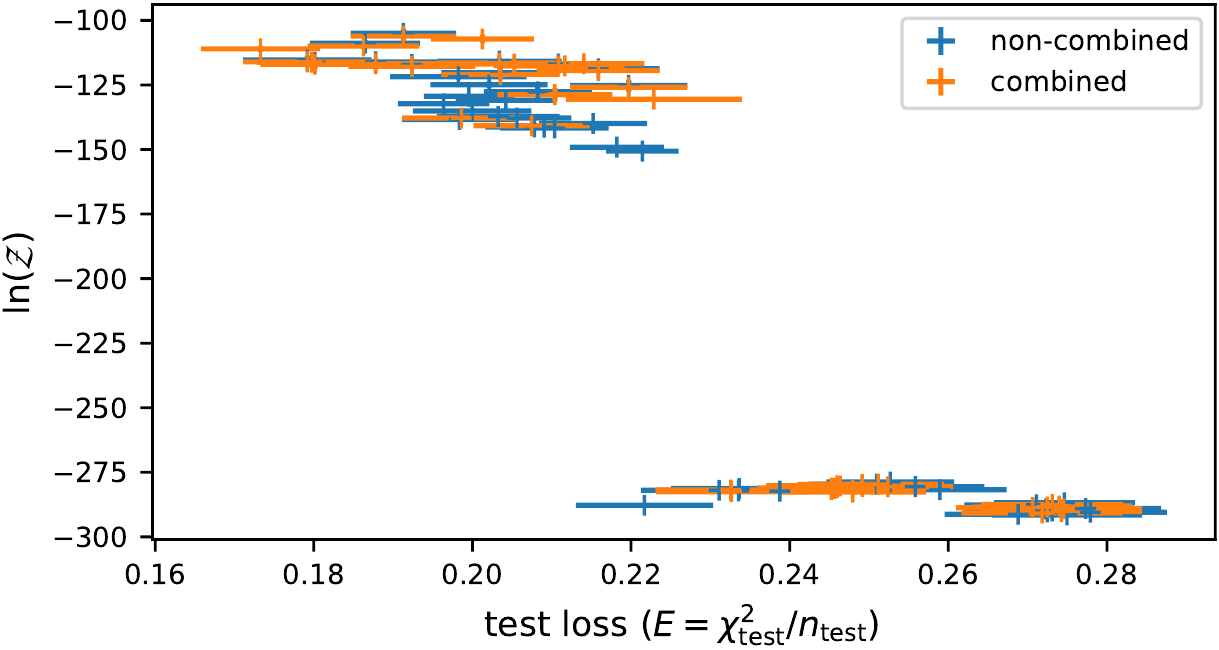}
  \caption{Log Bayesian evidence versus test loss values averaged over the ten different data randomisations, for all individual (blue) and ensembled (orange) BNNs.}
\label{f:all_mean_results}
  \end{center}
\end{figure}

The $\z$-test set performance trends were very similar to the results of the models which the respective combined runs comprised of, as can be seen in \cref{f:all_mean_results} which shows the $\z$ and test set losses for both individual and combined models. The evidences of the combined runs generally lie in the middle of values of the individual ones which is to be expected, since the combined evidences are just a linear combination of the evidences of the individual runs, weighted by their priors. Test set performance was also very similar on average, and so the separation in the $\z$--test loss plane corresponding to varying and fixed hyperparameter models persist with the combined models. Though the performances were on average very similar, as mentioned in the main text the lowest test loss was obtained using a combined model; the ensemble which combined all models trained with layer granularity hyperparameters achieved a test loss of 0.1732, while the best individual model was the $(l_{1},l_{2}) = (4,4)$ model with layer hyperparameter granularity which obtained a test loss value of 0.1791.

\begin{figure}
  \begin{center}
    \includegraphics{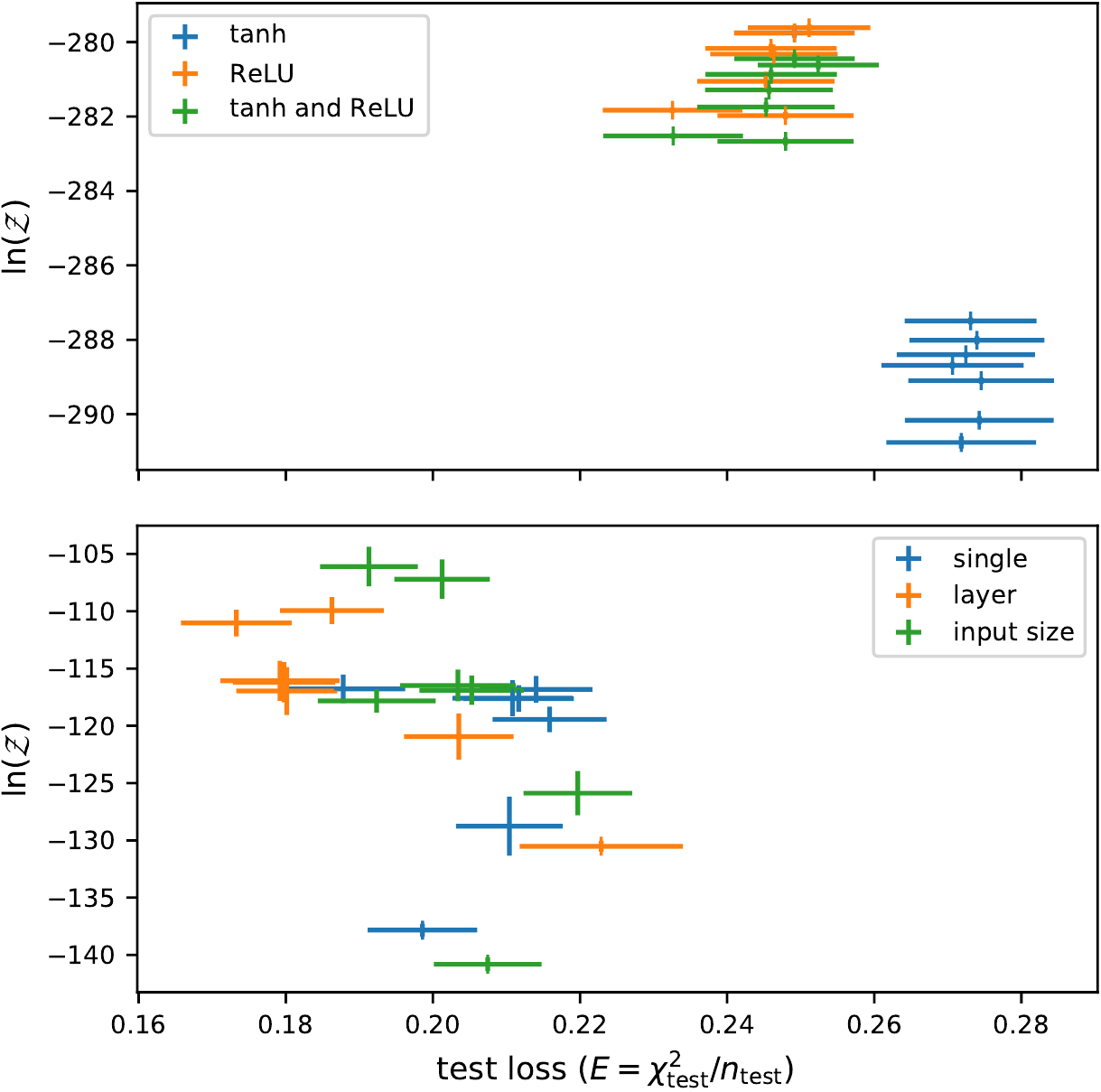}
  \caption{Top plot: Log Bayesian evidence versus test loss values averaged over the ten different data randomisations, for all the $\tanh$ activation function BNNs combined (blue points), the $\relu$ activation function BNNs combined (orange points), and combinations of BNNs with both types of activation function (green points). For each of these, the seven combinations (datapoints) are as follows: all one layer models combined, all two layer models combined, all three layer models combined, all four layer models combined, all two node-wide models combined, all four node-wide models combined, and all models combined. Bottom plot: Same as above but for single granularity random variable prior hyperparameter BNNs combined (blue points), layer granularity random variable prior hyperparameter BNNs combined combined (orange points), and input size granularity random variable prior hyperparameter BNNs combined (green points).}
\label{f:comb_tanh_relu_aa_mean_results_sh_lh_ih_sv_mean_results}
  \end{center}
\end{figure}

\cref{f:comb_tanh_relu_aa_mean_results_sh_lh_ih_sv_mean_results} compares various combined models. The top plot shows all models with fixed prior hyperparameters, while the bottom plot compares models with all three different granularities of hyperparameters. The different ensembles considered for each type are listed in \cref{t:combined_averages}.

\subsection{Comparison traditional network training}
\label{s:tnnresults}

\begin{table}
    \centerline{
\begin{tabular}{lcccccccccc}
\toprule
Architectures: &  (2) &  (4) & (8) &  (2, 2) &  (4, 4) &  (2, 2, 2) &  (4, 4, 4) &  (2, 2, 2, 2) &  (4, 4, 4, 4) \\
\midrule
$\tanh$ loss: &     0.2892 &     0.2942 &     0.2374 &       0.2776 &       0.2511 &         0.2884 &         0.2284 &           0.3085 &           0.2363 \\
$\relu$ loss: &     0.4025 &     0.2782 &     0.2609 &       0.5472 &       0.2957 &         0.7637 &         0.2877 &           0.7375 &           0.3068 \\
\bottomrule
\end{tabular}
}
\caption{Test losses obtained from networks trained with traditional maximum likelihood, backward propagation optimisation techniques.
\label{t:tnn_test_losses}}
\end{table}

As a means of providing a baseline for the previous analyses, we trained the same network architectures (with $\tanh$ or $\relu$ activations) using a basic back propagation methods \citep{rumelhart1985learning, mackay2003information} in their simplest form (i.e. no regularisation, no hyperparameter tuning). We refer to these networks as traditional neural networks (TNNs). The networks were trained for 1000 epochs each using the Adam optimisation algorithm \citep{kingma2014adam}, on the same training/test data splits as considered for the BNNs. Once again we used 10 different data randomisations and averaged the results. We note in passing that the maximum likelihood parameters found by the two different methods of training the networks showed no correlation, suggesting the optimisation and sampling methods are exploring the parameter spaces in quite different ways. For networks with $\tanh$ activations, six of the TNN test set estimates estimates were inferior to the no variable hyperparameter BNN estimates, while four were superior. When variable hyperparameters were used, all 10 BNNs performed better than the TNN equivalent architectures. For $\relu$ networks the simplest BNNs outperformed the TNNs in all 10 cases, and the $\tanh$ TNNs outperformed the $\relu$ TNNs in all but one case, again emphasising the significance of the $\relu$ BNNs outperforming their $\tanh$ equivalents. \cref{t:tnn_test_losses} summarises the performance of the TNNs.

\end{document}